\documentclass{article}

    \PassOptionsToPackage{numbers, compress}{natbib}
 \usepackage[preprint]{neurips_2025}

\usepackage[utf8]{inputenc} 
\usepackage[colorlinks=true]{hyperref}       
\usepackage{url}            
\usepackage{booktabs}       
\usepackage{amsfonts}       
\usepackage{nicefrac}       
\usepackage{microtype}      
\usepackage[table]{xcolor}  

\definecolor{mydeepblue}{RGB}{39, 60, 117}
\definecolor{myblue}{RGB}{200, 214, 229}
\hypersetup{linkcolor=mydeepblue,citecolor=mydeepblue,urlcolor=mydeepblue}  

\definecolor{aheNavy}{HTML}{1B262C}
\definecolor{aheBlue}{HTML}{0F4C75}
\definecolor{aheTeal}{HTML}{3282B8}
\definecolor{aheSky}{HTML}{BBE1FA}
\usepackage{amsmath}
\usepackage{cleveref}
\usepackage[most]{tcolorbox} 
\usepackage{wrapfig}        
\usepackage{algorithm}      
\usepackage{algpseudocode}  
\usepackage[T1]{fontenc} 
\usepackage{listings}       
\usepackage{pifont}         
\usepackage{placeins}       
\usepackage{eso-pic}        
\newcommand\blfootnote[1]{%
  \begingroup
  \renewcommand\thefootnote{}%
  \footnotetext{#1}%
  \endgroup
  \addtocounter{footnote}{-1}%
}

\usepackage{fontawesome5}

\title{Agentic Harness Engineering: Observability-Driven Automatic Evolution of Coding-Agent Harnesses}

\author{%
 \textbf{Jiahang Lin\textsuperscript{1,4}$^{*\ddagger}$},
 \textbf{Shichun Liu\textsuperscript{1}$^{*\ddagger}$},
 \textbf{Chengjun Pan\textsuperscript{2}$^{*\ddagger}$},
 \textbf{Lizhi Lin\textsuperscript{3}},
 \textbf{Shihan Dou\textsuperscript{1}}, \\
 \textbf{Zhiheng Xi\textsuperscript{1}},
 \textbf{Xuanjing Huang\textsuperscript{1}},
 \textbf{Hang Yan\textsuperscript{3}},
 \textbf{Zhenhua Han\textsuperscript{3}$^\dagger$},
 \textbf{Tao Gui\textsuperscript{1,4}$^\dagger$},
 \textbf{Yu-Gang Jiang\textsuperscript{1}}
 \\[2pt]
 \textsuperscript{1}Fudan University \quad
 \textsuperscript{2}Peking University  \\
 \textsuperscript{3}Shanghai Qiji Zhifeng Co., Ltd \quad
 \textsuperscript{4}Shanghai Innovation Institute \\
 \faGithub~\href{https://github.com/china-qijizhifeng/agentic-harness-engineering}{\texttt{china-qijizhifeng/agentic-harness-engineering}}
}

\begin{document}

\AddToShipoutPictureBG*{%
  \AtPageUpperLeft{%
    \put(110,-75){\includegraphics[height=1.2cm]{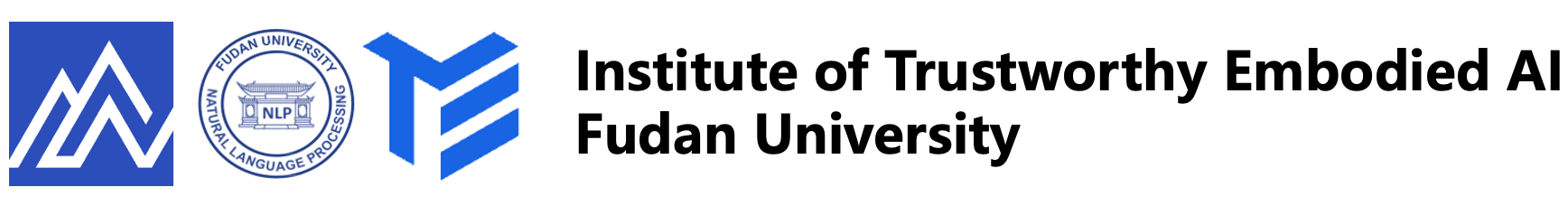}}
  }%
}

\maketitle
\blfootnote{$^*$Equal contributions.\  $^\dagger$Corresponding authors: \url{hzhua201@gmail.com}, \url{tgui@fudan.edu.cn}.\ \ $^\ddagger$Work done during an internship at Shanghai Qiji Zhifeng Co., Ltd. 
}

\begin{abstract}
Harnesses are now central to agent performance, mediating how models interact with tools and execution environments.
Yet harness engineering remains a manual craft, because automating it faces a heterogeneous action space across editable components, voluminous trajectories that bury actionable signal, and edits whose effect is hard to attribute.
We introduce \textbf{A}gentic \textbf{H}arness \textbf{E}ngineering (\textbf{AHE}), a closed loop that addresses these challenges through three matched observability pillars: \ding{182}~\emph{component observability} gives every editable harness component a file-level representation so the action space is explicit and revertible; \ding{183}~\emph{experience observability} distills millions of raw trajectory tokens into a layered, drill-down evidence corpus that an evolving agent can actually consume; and \ding{184}~\emph{decision observability} pairs every edit with a self-declared prediction, later verified against the next round's task-level outcomes.
Together, these pillars turn every edit into a falsifiable contract, so harness evolution proceeds autonomously without collapsing into trial-and-error.
Empirically, ten AHE iterations lift pass@1 on Terminal-Bench 2 from 69.7\% to 77.0\%, surpassing the human-designed harness Codex (71.9\%) and the self-evolving baselines ACE and Training-Free GRPO.
The frozen harness transfers without re-evolution: on SWE-bench-verified it achieves the highest aggregate success while using $12\%$ fewer tokens than the seed, and on Terminal-Bench 2 it yields $+5.1$ to $+10.1$\,pp cross-family gains across three alternate model families, indicating the evolved components encode general engineering experience rather than benchmark-specific tuning.
Ablations further localize the gain to tools, middleware, and long-term memory rather than the system prompt.
These results position observability-driven evolution as a practical pathway to keep coding-agent harnesses continually improving alongside their base models.
\end{abstract}

\begin{figure}[h]
    \centering
    \includegraphics[width=0.95\linewidth]{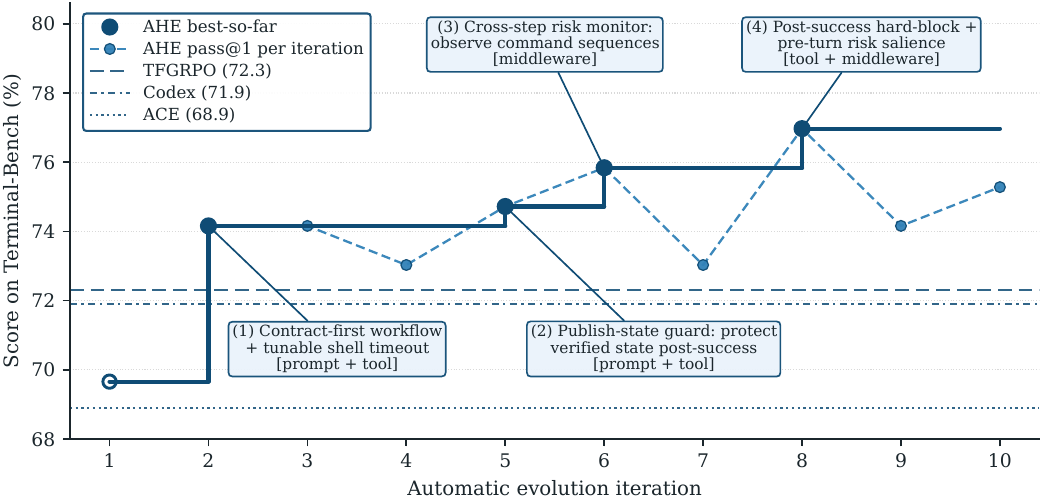}
    \caption{\textbf{AHE evolves a bash-only seed past every human-designed and self-evolving baseline on Terminal-Bench 2.} All three role agents share one base model, isolating the gain to harness edits rather than analyzer or editor capability.}
    \label{fig:training_curve}
\end{figure}

\section{Introduction}
\label{sec:introduction}

Coding agents are increasingly deployed on long-horizon software-engineering tasks, with measurable progress on issue resolution over real-world code repositories~\citep{jimenezSWEbenchCanLanguage2023,yangSWEbenchMultimodalAI2024,dengSWEBenchProCan2025} and multi-step terminal workflows~\citep{merrillTerminalBenchBenchmarkingAgents2026a}.
In practice, such progress relies not only on the underlying language model, but equally on the surrounding engineering components: the system prompt that shapes work style, the tools that expose the file system and shell, and the middleware that controls context, execution, and recovery.
This collection of model-external, editable components is collectively referred to as the agent's \emph{harness}~\citep{rajasekaranHarnessDesignLongrunning2026,lopopoloHarnessEngineeringLeveraging2026,wangOpenHandsOpenPlatform2025,yangSWEagentAgentComputerInterfaces2024,steinbergerOpenClawPersonalAI2026,HermesAgentAgent}.


Harness design materially shifts task completion on long-horizon coding benchmarks, even with the base model held fixed~\citep{trivedyImprovingDeepAgents2026,wangOpenHandsOpenPlatform2025}, making harness engineering a first-class lever for improving coding agents.
Moreover, the optimal harness is model-specific: a harness tuned for one base model often underperforms on another and must be re-adapted as the base model changes.
In current practice, this adaptation is performed manually—developers inspect trajectories, identify recurring failure patterns, and hand-craft edits across prompts, tools, middleware, and skills.
Yet as base models advance rapidly~\citep{xiaomimimoteamMiMoV25Pro2026,teamQwen36PlusRealWorld2026,yangQwen3TechnicalReport2025a,DeepSeek_V4pdfDeepseekaiDeepSeekV4Pro2026,kimiteamKimiK26Tech2026,teamKimiK25Visual2026}, this manual loop struggles to keep pace, creating a widening gap between model capability and the harness needed to realize it~\citep{steinbergerOpenClawPersonalAI2026}.


An intuitive direction is to automate this loop with an \textit{evolution agent} that optimizes harness components based on experience~\citep{agrawalGEPAReflectivePrompt2025a,zhangAgenticContextEngineering2025a,caiTrainingFreeGroupRelative2025}.
However, few existing approaches jointly evolve the full set of editable components~\citep{leeMetaHarnessEndtoEndOptimization2026}; most focus on a single component, typically the prompt~\citep{shinnReflexionLanguageAgents2023c,zhaoExpeLLLMAgents2024,madaanSelfRefineIterativeRefinement2023c}, skills~\citep{maSkillClawLetSkills2026b,xiaSkillRLEvolvingAgents2026c}, or an in-context playbook~\citep{zhangAgenticContextEngineering2025a}.
Jointly evolving multiple components end-to-end faces two structural obstacles: long, unstructured trajectories yield little actionable signal, and tightly coupled harness frameworks make edits beyond the prompt error-prone.
This leaves the central question of agent-driven harness evolution open:
\textit{How can an evolution agent \textbf{jointly} and \textbf{stably} evolve all editable components of a coding agent's harness?}

\begin{figure}[t]
    \centering
    \includegraphics[width=0.8\linewidth]{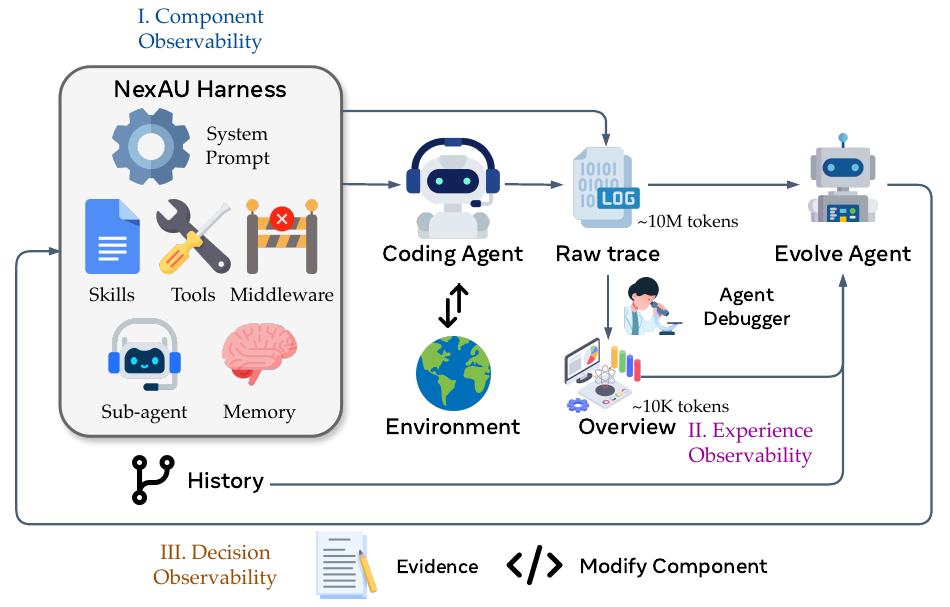}
    \caption{\textbf{The AHE pipeline links three observable surfaces into one closed loop.} Components, rollout experience, and edit decisions each surface as structured artifacts another agent reads, and every edit becomes a falsifiable prediction the next round verifies.}
    \label{fig:method}
\end{figure}

Our central insight is that this question is bottlenecked by \emph{observability}, not by agent capability: once the evolution agent receives structured context over a clear action space, it can reliably converge on better harness designs~\citep{suttonBitterLesson2019,zunicBitterLessonAgent2026}.
We implement this in \textbf{A}gentic \textbf{H}arness \textbf{E}ngineering 
(\textbf{AHE}, Figure~\ref{fig:method}), a closed loop driven by three observability pillars:
\ding{182}~\emph{component observability} via a decoupled harness that exposes seven editable component types as files, so each failure pattern maps cleanly to a single component class;
\ding{183}~\emph{experience observability} via a layered, drill-down evidence corpus distilled from millions of raw trajectory tokens, so the evolver consumes structured root causes rather than raw logs; and
\ding{184}~\emph{decision observability} via a change manifest that pairs every edit with a self-declared prediction, later verified against the next round's task-level outcomes, so each edit becomes a falsifiable contract and ineffective ones are reverted at file granularity.


We empirically validate AHE on Terminal-Bench 2\citep{merrillTerminalBenchBenchmarkingAgents2026a}: ten iterations lift pass@1 from 69.7\% to 77.0\%, surpassing the human-designed Codex~\citep{openaiCodexCLI2025} and the self-evolving baselines ACE~\citep{zhangAgenticContextEngineering2025a} and Training-Free GRPO~\citep{caiTrainingFreeGroupRelative2025}.
Without further evolution, the frozen harness transfers to SWE-bench-verified~\citep{jimenezSWEbenchCanLanguage2023}, and across three alternate base-model families it yields consistent pass@1 gains of $+5.1$ to $+10.1$ pp, with the largest on bases further from saturation, suggesting that AHE encodes coordination patterns that less-saturated models lean on more heavily.
A component ablation pinpoints where this gain lives: tools, middleware, and long-term memory each carry the improvement on their own, while the system prompt alone regresses, indicating that factual harness structure transfers across tasks and models whereas prose-level strategy does not.

This paper makes three contributions:
\begin{itemize}
    \item We formulate \emph{agent-driven harness evolution} for coding agents and propose \textbf{AHE}, which identifies \emph{observability across components, trajectories, and decisions} as the design pivot and turns every harness edit into a falsifiable, file-level contract through three observability pillars: a decoupled component substrate, a layered trajectory-distillation pipeline, and a change manifest whose self-declared predictions are verified by next-round task deltas.
    \item We empirically show that AHE lifts pass@1 on Terminal-Bench~2 from 69.7\% to 77.0\%, surpasses human-designed and automated baselines, and produces a frozen harness that transfers across benchmarks and base-model families.
    \item Our analysis reveals two limits of agent-driven evolution: harness components interact non-additively, so stacking effective edits caps the aggregate gain; and the loop's self-attribution is reliable for fixes but blind to regressions, pinpointing regression foresight as the clearest direction for future self-evolution loops.
\end{itemize}

\section{Related Work}
\label{sec:related}

\subsection{Harness Engineering and Evaluation for Coding Agents}
\label{ssec:related_harness}
Harness engineering refers to the practice of designing the system surrounding the model, including its tools, interfaces, memory, execution constraints, and feedback loops, which together shape what an agent can do on long-horizon tasks~\citep{rajasekaranHarnessDesignLongrunning2026,lopopoloHarnessEngineeringLeveraging2026,trivedyImprovingDeepAgents2026,anthropicClaudecode2025,steinbergerOpenClawPersonalAI2026,HermesAgentAgent}.
Concretely, the harness mediates how the model perceives and acts on its environment: it exposes the action and observation interfaces over which tool-augmented reasoning unfolds~\citep{anthropicClaudecode2025}, custom agent-computer interfaces for repository navigation, file editing, and command execution~\citep{yangSWEagentAgentComputerInterfaces2024}, as well as sandboxed execution and orchestration support that keep long-horizon runs reproducible~\citep{wangOpenHandsOpenPlatform2025}.

Verifying that such systems actually help has driven the parallel maturation of coding-agent evaluation along two axes: task horizon and environmental realism.
Coverage extends from short-horizon function-level benchmarks focused on contamination and freshness control~\citep{zhuoBigCodeBenchBenchmarkingCode2024,jainLiveCodeBenchHolisticContamination2024}, through repository-scale executable patch resolution~\citep{jimenezSWEbenchCanLanguage2023,yangSWEbenchMultimodalAI2024,dengSWEBenchProCan2025}, to multi-hour, terminal-driven workflows that exercise long-horizon, realistic execution~\citep{miserendinoSWELancerCanFrontier2025,chanMLEbenchEvaluatingMachine2024,merrillTerminalBenchBenchmarkingAgents2026a}.
A parallel infrastructure track packages executable runtimes and verifiers around these benchmarks~\citep{panTrainingSoftwareEngineering2025,jainR2EGymProceduralEnvironment2025,zengSWEHubUnifiedProduction2026}, whose attention to reproducible, traceable, and verifiable execution directly motivates the observation system AHE builds on.

\subsection{Automated Optimization of LLM Agents}
\label{ssec:related_auto}

Approaches to automated agent optimization differ in what evidence the optimizer observes and what it can edit.
Some revise the agent's own outputs through episodic critique and reflection~\citep{madaanSelfRefineIterativeRefinement2023c,shinnReflexionLanguageAgents2023c,guoCritiQMiningData2025}.
Others target prompts and instructions~\citep{khattabDSPyCompilingDeclarative2023}: structured playbooks~\citep{zhangAgenticContextEngineering2025a}, semantic-advantage priors~\citep{caiTrainingFreeGroupRelative2025}, jointly optimized instruction-demonstration pipelines for multi-stage programs~\citep{opsahl-ongOptimizingInstructionsDemonstrations2024a}, and reflective updates driven by Pareto-frontier traces~\citep{agrawalGEPAReflectivePrompt2025a}.
A separate line edits program structure itself, in the form of skill libraries~\citep{wangVoyagerOpenEndedEmbodied2023a}, scored program and agent archives evolved through mutation~\citep{novikovAlphaEvolveCodingAgent2025a,huAutomatedDesignAgentic2024}, and graph-structured workflows searched or learned from rollouts~\citep{zhangAFlowAutomatingAgentic2024,zhouSymbolicLearningEnables2024a}.

AHE tunes the full harness as a combinatorial whole rather than a single editable surface, so cross-component trade-offs become legible to the optimizer.
It also keeps the human prior minimal, leaving methodology for the optimizer to discover from rollouts rather than fixing it by hand.
We describe the substrate, trajectory analysis, and iteration that realize these choices in Section~\ref{sec:method}.

\section{Method}
\label{sec:method}

AHE turns harness optimization into a closed loop driven by another agent, with the base model held fixed and only the explicit harness edited.
Our design principle is that every phase of this loop must be \emph{observable}: AHE faithfully records the artifacts each phase produces (the harness components an iteration writes, the rollout trajectories it generates, the edit decisions it commits) and represents them in structured, layered forms that another agent can read and act on.

Three observability layers implement this principle.
\textbf{Component observability} (\S\ref{sec:method:harness}) is realized by a decoupled, file-level harness substrate that maps each failure pattern to a single component class.
\textbf{Experience observability} (\S\ref{sec:method:adb}) is realized by a layered evidence corpus distilled from raw rollouts and indexed for drill-down access.
\textbf{Decision observability} (\S\ref{sec:method:evolve}) is realized by a change manifest that pairs every edit with a self-declared prediction the next round verifies.
The three layers compose into the iteration of Algorithm~\ref{alg:ahe}, which runs unattended round after round.

\subsection{NexAU: an editable, decoupled harness substrate}
\label{sec:method:harness}

We instantiate the harness $H$ on the NexAU framework~\citep{nex-agiNexAUAUAgent2025,teamNexN1AgenticModels2025}, which exposes seven orthogonal component types as explicit files at fixed mount points in a single workspace: system prompt, tool description, tool implementation, middleware, skill, sub-agent configuration, and long-term memory.
The component types are loosely coupled, so adding a middleware does not require editing the system prompt, and adding a skill does not require touching any tool.

This decoupling is what realizes \textbf{component observability}: each failure pattern maps to a single component class, giving the evolve agent a clean action space and localizing every pass-rate change to one file rather than scattering it across hundreds of lines of unstructured prompt prose. Each logical edit becomes one commit on the workspace's git history, which yields file-level diffs and rollback granularity for free.

Our seed harness $H_0$ is deliberately minimal: a single shell-execution tool, no middleware, no skills, no sub-agents. A seed already fitted to the target benchmark would contaminate every subsequent edit's attribution, since we could not tell whether a gain came from the loop or from the seed. The minimal seed forces every component AHE adds to earn its place against measured rollouts.

\subsection{Agent Debugger: layered trajectory evidence}
\label{sec:method:adb}

We generate $k$ traces for each task in a benchmark using a harness $H$, which may contain errors resulting from the deficiencies of the harness that can be acted on, but scattered across millions of tokens of raw messages. To extract insights from agent trajectories and enable \textbf{experience observability}, we apply Agent Debugger~\citep{linAgentDebuggerUnderstanding2026} framework to use an agent to explore trajectories framed as a navigable, file-based environment where each trajectory message lives in its own file and is reached through generic shell and scripting tools. Traces with the same query are placed in one environment, and the debugger is required to analyze the root cause of the failure or the success pattern, which is stored in \emph{per-task analysis} report for each task. The analysis also includes pass/fail status of the task to ground the Evolve Agent. Finally, a \emph{benchmark-level overview} is aggregated from every report into a single document as an entry point for every iteration. 

In addition to these reports, we also provide \emph{original} traces in case the agents need to verify the claims in the reports. The traces are provided both in raw form and lightly processed to remove unnecessary content. All of these content is provided as files allowing \textit{progressive disclosure}~\citep{EffectiveContextEngineering} which saves on tokens and enable better agent decisions.

\subsection{Evolve Agent: evidence-driven, auditable edits}
\label{sec:method:evolve}

The Evolve Agent closes the AHE loop. In each round it reads the layered evidence corpus produced by the Agent Debugger, decides which harness components to add, modify, or remove, applies those edits to the workspace, and records the reasoning behind every edit. Two constraints govern these edits, and together they realize \textbf{decision observability}: every edit becomes a falsifiable, file-level claim recorded in a versioned manifest, and the next round's verdict either confirms or reverts it.

The first constraint is \textit{controllability}: the Evolve Agent writes only inside the harness workspace, while the runs directory, tracer, verifier, and LLM configuration are read-only, and the seed system prompt (Appendix~\ref{app:prompts:code-seed}) is marked non-deletable.
These restrictions block the shortcuts an unconstrained self-modifier would take, such as disabling the verifier, swapping the model, or raising the reasoning budget, and keep every recorded gain attributable to harness edits.

The second constraint is that every change is \textit{evidence-driven} and ships with a recorded prediction.
Each edit attaches a manifest entry that names the failure evidence, the inferred root cause, the targeted fix, and a predicted impact comprising both expected fixes and at-risk regressions; this manifest is the loop's evidence ledger (see Appendix~\ref{app:prompts:evolve}).
In the next round, the loop intersects the predicted-fix and predicted-regression sets with the observed task-level deltas to produce a per-edit verdict.
Each edit thereby becomes falsifiable by the next evaluation, which replaces rationale-driven self-justification with a measurable contract between rounds.

\begin{algorithm}[htbp]
\caption{AHE outer loop.}
\label{alg:ahe}
\begin{algorithmic}[1]
\Require seed harness $H_0$, base model $M$, benchmark $D$, rollouts per task $k$, max iterations $N$
\State $H_{\text{best}} \gets H_0$
\For{$t = 1$ to $N$}
    \State $T_t \gets \textsc{Rollout}(M, H_{t-1}, D, k)$ \Comment{phase 1: $k$ rollouts per task}
    \State $\widetilde{T}_t \gets \textsc{Clean}(T_t)$ \Comment{phase 2: normalize traces into canonical form}
    \If{$t \geq 2$} \Comment{phase 3: attribute prior manifest, then rollback}
        \State $V_t \gets \textsc{Attribute}(C_{t-1}, T_{t-1}, T_t)$
        \State $H_{t-1} \gets \textsc{Rollback}(H_{t-1}, V_t)$
    \Else
        \State $V_t \gets \emptyset$
    \EndIf
    \State $R_t \gets \textsc{AgentDebugger}(\widetilde{T}_t)$ \Comment{phase 4: layered distillation}
    \State $(H_t, C_t) \gets \textsc{Evolve}(H_{t-1}, R_t, V_t)$ \Comment{phase 5: workspace edits + new manifest}
    \State $\textsc{Commit}(H_t, C_t, t)$ \Comment{phase 6: tag iteration in git}
    \If{$\textsc{Pass@1}(T_t) > \textsc{Pass@1}(H_{\text{best}})$} $H_{\text{best}} \gets H_t$ \EndIf
\EndFor
\State \Return $H_{\text{best}}$
\end{algorithmic}
\end{algorithm}

Algorithm~\ref{alg:ahe} composes the three substrates into one iteration: rollout, clean, attribute the prior manifest and revert rejected edits, distill, edit, commit.
We run $k\geq 2$ rollouts per task so each task carries a pass-rate signal, which stabilizes pass@1 and lets partial-pass tasks anchor comparative diagnosis.
Attribution runs \emph{before} distillation, so its verdict lands inside the evidence corpus and binds each prior manifest entry as a contract rather than a rationale.
A one-shot explore agent (Appendix~\ref{app:prompts:explore}) runs in parallel with iteration~$1$ to seed a small number of reusable skills from the NexAU source and public coding-agent references.
These skills receive no special protection: from iteration~$2$ onward the Evolve Agent may keep, refine, or remove them based on observed rollouts.

\section{Experiments}
\label{sec:experiments}

We organize our empirical study around three questions: where AHE sits on the map of existing approaches to harness design, whether what it produces is portable beyond its optimization target, and what inside the loop drives the gain.

\begin{tcolorbox}[
    colback=blue!3,
    colframe=blue!40,
    arc=2pt,
    boxrule=0.5pt,
    left=6pt, right=6pt, top=4pt, bottom=4pt,
    title={\textbf{Research Questions}},
    fonttitle=\bfseries,
    coltitle=black,
    colbacktitle=blue!10
  ]
  \begin{enumerate}
    \setlength{\leftmargini}{1.5em}
    \setlength{\itemsep}{2pt}
    \setlength{\parsep}{0pt}
      \item \textbf{RQ1} (\S\ref{sec:experiments:main})\textbf{: Why agentic harness engineering, rather than human-engineered harnesses or other automated methods?}
      \item \textbf{RQ2} (\S\ref{sec:experiments:transfer})\textbf{: Does agentic harness engineering overfit to its optimization target?}
      \item \textbf{RQ3} (\S\ref{sec:experiments:mechanism})\textbf{: What inside AHE drives its gains, and how reliable is the loop's self-attribution?}
  \end{enumerate}
\end{tcolorbox}

\subsection{Setup}
\label{sec:experiments:setup}

\paragraph{Evaluation.}
\label{sec:experiments:metrics}
We drive evolution on the full 89 tasks of Terminal-Bench 2~\citep{merrillTerminalBenchBenchmarkingAgents2026a}, split as 4 easy, 55 medium, and 30 hard, with per-task timeout extended to 1 hour. For cross-benchmark transfer we evaluate the AHE harness on SWE-bench-verified~\citep{jimenezSWEbenchCanLanguage2023}, 500 tasks across seven repositories. 
We report two metrics per configuration: \textit{pass@1}, the mean binary success rate over $k$ rollouts per task; and \textit{tokens/trial}, the mean per-trial total of prompt plus completion tokens across all LLM calls, in thousands. Infrastructure-aborted or timed-out trials count as failures under pass@1 (matching the official terminal-bench leaderboard) and are excluded from token means to avoid truncated figures. Runtime infrastructure (framework, dispatcher, sandbox, tracer, and concurrency) is detailed in Appendix~\ref{app:setup}.

\paragraph{Models.}
For both the evolution loop and the main experiment of \S\ref{sec:experiments:main}, all three role agents (the Code Agent, the Agent Debugger, and the Evolve Agent) share one base model, GPT-5.4~\citep{openaiIntroducingGPT542026} at the high reasoning setting. 
For cross-model transfer (\S\ref{sec:experiments:transfer}), we re-evaluate the Code Agent on five alternate bases: GPT-5.4~\citep{openaiIntroducingGPT542026} at medium and xhigh reasoning, qwen-3.6-plus~\citep{teamQwen36PlusRealWorld2026,yangQwen3TechnicalReport2025a}, gemini-3.1-flash-lite-preview~\citep{googleGemini31FlashLiteModelCard2026}, and deepseek-v4-flash~\citep{DeepSeek_V4pdfDeepseekaiDeepSeekV4Pro2026}.

\subsection{RQ1: Main Results}
\label{sec:experiments:main}


\setlength{\intextsep}{2pt}
\setlength{\columnsep}{8pt}
\begin{wraptable}{r}{0.5\linewidth}
    \centering
    \footnotesize
    \setlength{\tabcolsep}{4pt}
    \renewcommand{\arraystretch}{1.0}
    \setlength{\abovecaptionskip}{2pt}
    \setlength{\belowcaptionskip}{2pt}
    \caption{Pass@1 on Terminal-Bench 2 across 89 tasks, by official difficulty. NexAU\textsubscript{0} is the shared seed; ACE, Training-Free GRPO, and \textbf{AHE} are three self-evolution loops layered on top of it. Bold marks the best per column; ties are all bold.}
    \label{tab:main-results}
    \begin{tabular}{lcccc}
        \toprule
        Method & All & Easy & Med. & Hard \\
               & {\scriptsize 89} & {\scriptsize 4} & {\scriptsize 55} & {\scriptsize 30} \\
        \midrule
        \rowcolor{gray!20} \multicolumn{5}{l}{\textbf{Human-designed harness}} \\
        OpenCode               & 47.2\% & 75.0\% & 52.7\% & 33.3\% \\
        Terminus-2             & 62.9\% & 75.0\% & 74.5\% & 40.0\% \\
        Codex              & 71.9\% & 75.0\% & 80.0\% & \textbf{56.7\%} \\
        \addlinespace
        \rowcolor{gray!20} \multicolumn{5}{l}{\textbf{Self-evolved from NexAU\textsubscript{0}}} \\
        NexAU\textsubscript{0} & 69.7\% & 87.5\% & 78.2\% & 51.7\% \\
        ACE                    & 68.9\% & 91.7\% & 78.2\% & 48.9\% \\
        TF-GRPO                & 72.3\% & \textbf{100.0\%} & 79.4\% & 55.6\% \\
        \textbf{AHE}           & \textbf{77.0\%} & \textbf{100.0\%} & \textbf{88.2\%} & 53.3\% \\
        \bottomrule
    \end{tabular}
\end{wraptable}

We run a single AHE campaign of ten iterations from the bash-only \textbf{NexAU\textsubscript{0}} seed (\S\ref{sec:method:harness}) on Terminal-Bench 2, finishing in roughly 32 hours; the best resulting configuration is reported as \textbf{AHE}. The two self-evolve baselines ACE~\citep{zhangAgenticContextEngineering2025a} and Training-Free GRPO (TF-GRPO)~\citep{caiTrainingFreeGroupRelative2025} start from the same NexAU\textsubscript{0} seed.

\paragraph{AHE outperforms both human-designed and self-evolve baselines.}
AHE outperforms every baseline on our panel: three human-designed harnesses, OpenCode~\citep{anomalyOpencodeOpenSource2025}, Terminus-2~\citep{harborTerminus22026}, and Codex~\citep{openaiCodexCLI2025}, and the two self-evolve baselines.
Figure~\ref{fig:training_curve} shows the gain accumulates across iterations, with continued evolution pushing pass@1 further above the NexAU\textsubscript{0} seed.
By difficulty, the only exception is the Hard tier, where AHE marginally trails Codex. We trace this gap to interference between AHE's components on long-horizon tasks rather than to a missing capability: swapping AHE's long-term memory alone into the NexAU\textsubscript{0} seed, without the other AHE components, already surpasses Codex on Hard (\S\ref{sec:experiments:rq3a}).

\paragraph{Prompt-only self-evolution misses the components that carry AHE's gain.}
The gaps to ACE and TF-GRPO trace to a layer mismatch. ACE distills natural-language playbooks the agent reads in-context, and TF-GRPO is a trajectory-feedback variant of GRPO that reinforces successful tool sequences; neither method opens the surrounding scaffolding to edits.
AHE instead jointly evolves system prompt, tools, middleware, and long-term memory, and \S\ref{sec:experiments:rq3a} shows the gain concentrates in the latter three components, exactly the layers ACE and TF-GRPO leave untouched.

\subsection{RQ2: Transfer to Unseen Tasks and Base Models}
\label{sec:experiments:transfer}

AHE's harness is evolved on Terminal-Bench 2 with GPT-5.4 high. We probe whether it encodes general coding-agent experience or overfits to that target by transferring the evolved harness unchanged, without further evolution, to two off-target settings: a different task surface (SWE-bench-verified) and three alternate base models.

\begin{table}[t]
    \centering
    \small
    \setlength{\tabcolsep}{5pt}
    \renewcommand{\arraystretch}{1.05}
    \caption{Cross-benchmark transfer on SWE-bench-verified. ACE, Training-Free GRPO (TF-GRPO), and \textbf{AHE} share the \textbf{NexAU\textsubscript{0}} seed and differ only in their self-evolution loop; all four columns run on GPT-5.4. AHE and the two self-evolve baselines are evolved on Terminal-Bench 2 and evaluated without in-domain re-evolution. Per-column bold marks the best; ties are all bold.}
    \label{tab:swe-verified-analysis}
    \begin{tabular}{lc @{\hspace{10pt}} cccc @{\hspace{10pt}} cccc}
        \toprule
         &  & \multicolumn{4}{c}{Success rate $\uparrow$} & \multicolumn{4}{c}{Tokens k $\downarrow$} \\
        \cmidrule(lr){3-6} \cmidrule(lr){7-10}
        Repo & $N$
        & ACE & TF-GRPO & NexAU\textsubscript{0} & \textbf{AHE}
        & ACE & TF-GRPO & NexAU\textsubscript{0} & \textbf{AHE} \\
        \midrule
        All            & 500 & 74.6\% & 74.2\% & 75.2\% & \textbf{75.6\%} & 679 & 582 & 526 & \textbf{461} \\
        \midrule
        django         & 231 & 79.2\% & 78.8\% & 79.2\% & \textbf{81.0\%} & 707 & 583 & 527 & \textbf{484} \\
        sympy          &  75 & 69.3\% & 68.0\% & \textbf{70.7\%} & \textbf{70.7\%} & 602 & 572 & 494 & \textbf{479} \\
        sphinx-doc     &  44 & 61.4\% & 65.9\% & 68.2\% & \textbf{70.5\%} & 990 & 848 & 731 & \textbf{656} \\
        matplotlib     &  34 & 70.6\% & 70.6\% & \textbf{73.5\%} & \textbf{73.5\%} & 622 & 530 & 486 & \textbf{391} \\
        scikit-learn   &  32 & \textbf{93.8\%} & \textbf{93.8\%} & \textbf{93.8\%} & 87.5\% & 451 & 378 & 307 & \textbf{257} \\
        pydata         &  22 & \textbf{77.3\%} & \textbf{77.3\%} & \textbf{77.3\%} & 72.7\% & 563 & 516 & 386 & \textbf{338} \\
        astropy        &  22 & \textbf{59.1\%} & \textbf{59.1\%} & 54.5\% & 50.0\% & 546 & 470 & 667 & \textbf{277} \\
        \bottomrule
    \end{tabular}
\end{table}

\paragraph{Cross-benchmark transfer.}
We evaluate the AHE harness on SWE-bench-verified against the seed and the two self-evolve baselines (NexAU\textsubscript{0}, ACE, TF-GRPO) under identical infrastructure (Table~\ref{tab:swe-verified-analysis}).

ACE and TF-GRPO both regress below the NexAU\textsubscript{0} seed in aggregate success while spending $11\%$ to $29\%$ more tokens than the seed: the playbook ACE injects and the trajectory distribution TF-GRPO reinforces were distilled on Terminal-Bench 2 traces and ride the prompt at every model call, so on a different task surface that text adds cost without reshaping the underlying policy.

AHE instead achieves the highest aggregate, with the seed-relative gain concentrating on django and sphinx-doc, the two largest and most token-expensive repositories whose multi-step edit-and-verify loop matches the structure AHE's tools, middleware, and long-term memory compress on Terminal-Bench 2.
Marginal regressions appear only on the three smallest repositories, consistent with pass@1 variance on small repos exceeding the per-repo gain.
AHE also cuts aggregate tokens by $32\%$ against ACE, $21\%$ against TF-GRPO, and $12\%$ against the seed: encoding behavior in tools, middleware, and memory avoids the per-call re-derivation cost prompt-only baselines incur.

\begin{figure}[t]
    \centering
    \includegraphics[width=0.9\linewidth]{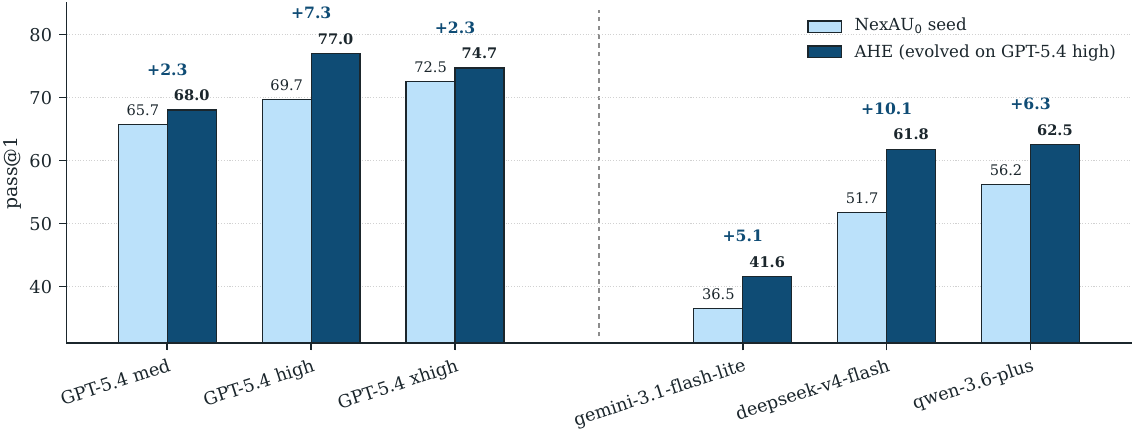}
    \caption{Cross-model transfer on Terminal-Bench 2, 89 tasks. The AHE workspace evolved on GPT-5.4 high is re-evaluated on each base without further evolution, paired against the NexAU\textsubscript{0} seed on the same base. All bases gain over the seed, with cross-family bases gaining more than within-family ones.}
    \label{fig:transfer-model}
\end{figure}

\paragraph{Cross-model transfer.}
We re-evaluate both the NexAU\textsubscript{0} seed and AHE on the five alternate bases listed in \S\ref{sec:experiments:setup}. Figure~\ref{fig:transfer-model} reports five positive pass@1 gains from $+2.3$ to $+10.1$\,pp.

Cross-family gains dominate within-family ones: deepseek-v4-flash moves $+10.1$\,pp from $51.7\%$ to $61.8\%$, qwen-3.6-plus $+6.3$\,pp from $56.2\%$ to $62.5\%$, and gemini-3.1-flash-lite-preview $+5.1$\,pp from $36.5\%$ to $41.6\%$, all above the $+2.3$\,pp on GPT-5.4 medium and xhigh. We read this as bases further from saturation leaning more on the coordination patterns AHE has fixed inside tools, middleware, and long-term memory, while a stronger base re-derives the same coordination from its prompt at low marginal cost.

Within one family the profile is non-monotone: $+2.3$\,pp on medium, $+7.3$\,pp on high from \S\ref{sec:experiments:main}, and $+2.3$\,pp on xhigh. AHE's step budget and per-task timeout were fitted to GPT-5.4 high during evolution; medium has more time-per-step slack but loses a reasoning tier of raw capability, while xhigh pushes more trials past the per-task timeout, which our pass@1 convention (\S\ref{sec:experiments:metrics}) counts as failures. Either direction discounts the gain.

The headline finding is that all five gains land positive: the AHE workspace is not specific to one provider's idioms or one reasoning depth. Their magnitude tracks the evolution operating point rather than raw base capability, so we treat the timeout-budget coupling as a generalization hazard discussed in our \nameref{sec:limitations} section.

\subsection{RQ3: Analysis}
\label{sec:experiments:mechanism}

We analyze the loop along two architectural choices that \S\ref{sec:method} places weight on: decomposed components (\S\ref{sec:experiments:rq3a}) and self-declared attribution (\S\ref{sec:experiments:rq3b}).

\subsubsection{RQ3a: where value accumulates across components}
\label{sec:experiments:rq3a}

Table~\ref{tab:ablations} decomposes the AHE gain at the component level by isolating one of the four evolved layers, long-term memory, tools, middleware, or system prompt, while holding the other three at their seed defaults. Three of the four single-component variants outperform the seed; the system-prompt swap is the only regression.

\begin{table}[htbp]
    \centering
    \small
     \vspace{\baselineskip}
    \caption{Component-level ablations on Terminal-Bench 2. Each ``+\,X only'' row swaps a single AHE component into the NexAU\textsubscript{0} seed; per-column best is bolded. Most single-component swaps already beat the seed, and the single-component gains overshoot full AHE's aggregate, indicating components interact non-additively rather than stacking cleanly.}
    \label{tab:ablations}
    \begin{tabular}{lcccc}
        \toprule
        Variant & All & Easy & Medium & Hard \\
                & {\scriptsize 89 tasks} & {\scriptsize 4 tasks} & {\scriptsize 55 tasks} & {\scriptsize 30 tasks} \\
        \midrule
        NexAU\textsubscript{0} & 69.7\% & 87.5\% & 78.2\% & 51.7\% \\
        \addlinespace
        + memory only         & 75.3\% & 50.0\% & 83.6\% & \textbf{63.3\%} \\
        + tool only           & 73.0\% & 75.0\% & 87.3\% & 46.7\% \\
        + middleware only     & 71.9\% & \textbf{100.0\%} & 81.8\% & 50.0\% \\
        + system\_prompt only & 67.4\% & 75.0\% & 78.2\% & 46.7\% \\
        \addlinespace
        \textbf{AHE} full     & \textbf{77.0\%} & \textbf{100.0\%} & \textbf{88.2\%} & 53.3\% \\
        \bottomrule
    \end{tabular}
\end{table}

\paragraph{Each component owns a different failure surface.}
Memory adds 12 boundary-case lessons (performance margin, queued-over-limit cancellation, evaluator-style closure, source-packaging layout); on Hard the lessons lift it above full AHE, while on Easy they reduce to superfluous re-verification.
Tools become a 1364-line shell that auto-surfaces contract hints from files near each command; on Medium it lands within $0.9$\,pp of full AHE, while on Hard a built-in publish guard closes the loop too early.
Middleware adds a finish-hook that forces one evaluator-isomorphic closure check; on Easy it clears every task, while on Hard it inflates turn count.
The system prompt encodes 79 lines of universal discipline whose executability depends on the other three; inserted alone it scores $-2.3$\,pp aggregate.

\paragraph{Components interact non-additively, capping the aggregate gain.}
The three positive single-component gains sum to $+11.1$\,pp against full AHE's $+7.3$\,pp, and on Hard the memory-only variant exceeds full AHE: memory, middleware, and the system prompt all push toward the same closure-style verification, so stacking them spends turns on redundant re-checks within the long-horizon budget.
Since the evolve agent optimises an aggregate dominated by 55 Medium tasks, it converges to a Medium-heavy trade-off that returns part of the Hard memory effect, and we leave interaction-aware evolution to future work.

\subsubsection{RQ3b: how reliably the loop's self-attribution tracks reality}
\label{sec:experiments:rq3b}

Each evolution round, our evolve model produces a change manifest naming which Terminal-Bench 2 tasks it expects to fix in the next round and which it flags at risk of regression. We compare the round-$N{-}1$ prediction against the round-$N$ ground truth, computing standard precision and recall over the 89 tasks separately for fixes and regressions.

\begin{figure}[t]
    \centering
    \begin{minipage}{0.4\linewidth}
        \centering
        \includegraphics[width=\linewidth]{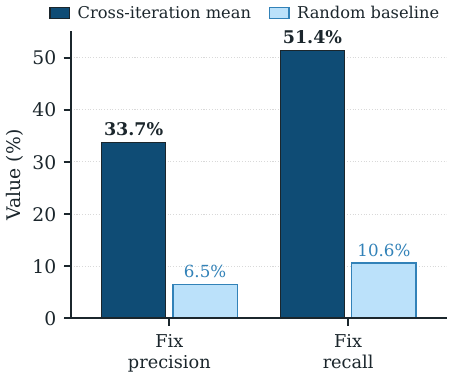}
    \end{minipage}
    \begin{minipage}{0.4\linewidth}
        \centering
        \includegraphics[width=\linewidth]{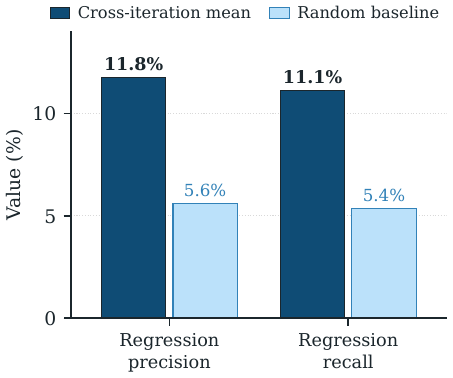}
    \end{minipage}
    \caption{Cross-iteration mean precision and recall of the evolve model's self-predictions across 9 evaluation rounds of the GPT-5.4 AHE loop on Terminal-Bench 2, alongside the random-prediction baseline. Left: fix predictions. Right: regression predictions.}
    \label{fig:pred-aggregate}
\end{figure}

\paragraph{Evidence-driven targeting.}
The fix panel of \Cref{fig:pred-aggregate} shows the evolve model's targeting is evidence-driven rather than guesswork. Cross-iteration fix-precision of \textbf{33.7\%} and fix-recall of \textbf{51.4\%} sit roughly 5x above the random-prediction baselines of 6.5\% and 10.6\%, so each harness edit lands on a real, agent-anticipated target rather than on an arbitrary subset of the panel.

\paragraph{Regression blindness.}
The regression panel tells the opposite story: cross-iteration regression-precision of \textbf{11.8\%} and regression-recall of \textbf{11.1\%} sit only about 2x above their random baselines of 5.6\% and 5.4\%, so most upcoming regressions go unforeseen. The agent can justify why an edit should help, but it cannot reliably name the tasks the same edit is about to break, which is what produces the non-monotone steps in the evolution curve of \S\ref{sec:experiments:main}. Closing this gap is the clearest direction for future self-evolution loops. Appendix~\ref{sec:appendix:self-attribution} gives the per-round breakdown.

\section{Conclusion}

We introduced \textbf{Agentic Harness Engineering (AHE)}, an observability-driven loop that turns a coding agent's harness into a learnable adaptation surface while the base model remains fixed. AHE exposes components as files, distills rollouts into a layered evidence corpus, and binds each edit to a falsifiable next-round prediction; ten iterations lift pass@1 on Terminal-Bench~2 from 69.7\% to 77.0\%, and the frozen harness transfers to SWE-bench-verified and three alternate model families. We see harness-level evolution as a complementary axis to model-side training: an externalized, auditable surface where coding-agent experience can accumulate.


\section*{Limitations}
\phantomsection
\label{sec:limitations}

This work studies a promising but high-variance setting, and the scope of our claims should be interpreted accordingly.

\paragraph{Benchmark scope.}
Our evaluation drives evolution on Terminal-Bench~2 and probes transfer on SWE-bench-verified. Even though the frozen harness transfers to a second task surface and to three alternate base-model families, broader programming languages, repository-scale deployments, and human-in-the-loop workflows remain untested.

\paragraph{Evolution operating point.}
AHE's step budget and per-task timeout were fitted to GPT-5.4 high during evolution, so cross-model transfer numbers conflate harness portability with operating-point coupling—within one family the gain is non-monotone across reasoning tiers (\S\ref{sec:experiments:transfer}). Untangling these factors will require re-running the loop under multiple operating points.

\paragraph{Self-modification governance.}
AHE bounds edits to a workspace, attributes every change in a versioned manifest, and rolls back ineffective edits at file granularity, but it does not provide a complete guardrail stack. Long-horizon harness cleanup and stronger misuse prevention remain incomplete, and AHE should be viewed as a controlled research prototype rather than a fully mature autonomous self-improvement system.

\bibliographystyle{plainnat}
\bibliography{ahe}


\appendix

\section{Experimental Setup: Full Details}
\label{app:setup}

This appendix expands the condensed Setup in \S\ref{sec:experiments:setup} with the formal metric definitions and the runtime infrastructure.

\paragraph{Seed agent.}
The seed configuration, denoted \textbf{NexAU\textsubscript{0}}, is a simple code agent built on the NexAU framework of \S\ref{sec:method:harness} that exposes only the \texttt{bash} tool to the model, with no skills, no middleware, and no long-term memory. Every iteration of the AHE outer loop edits this workspace, so all reported gains are measured against NexAU\textsubscript{0} as the common starting point.

\paragraph{Runtime infrastructure.}
All runs use the NexAU framework of \S\ref{sec:method:harness} to instantiate the coding agent. Harbor dispatches tasks, isolates each rollout, and verifies pass/fail, with a 3600-second per-task timeout. Every rollout runs inside a fresh E2B remote sandbox, so shell side-effects cannot leak between tasks. \texttt{InMemoryTracer} records trajectories and mirrors them to Langfuse. The Agent Debugger executes at concurrency 16 with a 600-second per-task timeout.

\paragraph{Hyperparameters.}
Table~\ref{tab:hparams} consolidates the operating points of the four agents and the outer loop, taken from the configuration snapshot of the reference run.

\begin{table}[h]
    \centering
    \small
    \setlength{\tabcolsep}{6pt}
    \renewcommand{\arraystretch}{1.05}
    \caption{Hyperparameters of the reference AHE run on Terminal-Bench~2, taken from the configuration snapshot of the run that produced the AHE numbers in Section~\ref{sec:experiments:main}. All four agents share GPT-5.4 as the backing model and differ only in reasoning tier, sampling, and per-component limits.}
    \label{tab:hparams}
    \begin{tabular}{lll}
        \toprule
        Block & Hyperparameter & Value \\
        \midrule
        Code Agent
            & Backing model              & GPT-5.4 \\
            & Reasoning effort           & high \\
            & Temperature                & 0.7 \\
            & Top-$p$                    & 0.95 \\
            & Max context                & 200{,}000 tokens \\
            & Max generation per turn    & 32{,}000 tokens \\
            & Max model turns            & 300 \\
        \midrule
        Agent Debugger
            & Backing model              & GPT-5.4 \\
            & Concurrent tasks           & 16 \\
            & Per-task timeout           & 600\,s \\
            & Retry attempts             & 3 \\
            & Retry backoff              & 2\,s \\
        \midrule
        Evolve Agent
            & Backing model              & GPT-5.4 \\
            & Reasoning effort           & xhigh \\
            & Temperature                & 0.7 \\
            & Max context                & 200{,}000 tokens \\
            & Max generation per turn    & 32{,}000 tokens \\
            & Max model turns            & 500 \\
            & Skill packages             & 3 \\
            & Context compaction threshold & 0.75 \\
        \midrule
        Explore Agent
            & Backing model              & GPT-5.4 \\
            & Reasoning effort           & xhigh \\
            & Wall-clock budget          & 60\,min \\
            & Web sources                & 10 \\
            & Code sources               & 1 \\
        \midrule
        Outer loop / dispatcher
            & Iterations $T$             & 10 \\
            & Rollouts per task $k$      & 2 \\
            & Concurrent rollouts        & 96 \\
            & Sandbox                    & E2B remote \\
            & Sandbox lifetime           & 3{,}600\,s \\
            & Dataset                    & Terminal-Bench~2, 89 tasks \\
        \bottomrule
    \end{tabular}
\end{table}

\paragraph{Terminal-bench difficulty labels.}
The official terminal-bench-2 leaderboard\footnote{\url{https://www.tbench.ai/benchmarks/terminal-bench-2}} partitions the 89-task subset into 4 easy, 55 medium, and 30 hard tasks.

\paragraph{pass@1.}
For a configuration on a task set $D$ with $k$ rollouts per task, let $r_{i,j} \in \{0, 1\}$ denote the binary reward of rollout $j$ on task $i$. The pass@1 score is the mean
\begin{equation}
    \mathrm{pass@1} = \frac{1}{k|D|} \sum_{i=1}^{|D|} \sum_{j=1}^{k} r_{i,j}.
\end{equation}
Trials that terminate on an infrastructure exception, such as a sandbox crash or API timeout, contribute $r=0$ rather than being dropped, a strictly harsher convention than discarding failures that keeps our numbers comparable to the official terminal-bench leaderboard. The rollout count $k$ varies across experiments; each table states it explicitly.

\paragraph{Token cost and Succ/Mtok.}
For token cost we count every LLM call as prompt plus completion across the rollout and report the mean over completed trials in thousands, denoted Tokens k; infrastructure-aborted trials are excluded to avoid truncated figures. To compare configurations that trade accuracy for cost we combine the two via
\begin{equation}
    \mathrm{Succ/Mtok} = \frac{\mathrm{pass@1} \times 10^{6}}{\mathrm{mean\ tokens\ per\ trial}},
\end{equation}
the expected number of successes per million tokens. The main paper reports pass@1 and Tokens k separately so each axis stays legible; Table~\ref{tab:swe-verified-succ-mtok} folds them into Succ/Mtok per repository on SWE-bench-verified, derived from the pass@1 and Tokens k columns of Table~\ref{tab:swe-verified-analysis}.

\begin{table}[t]
    \centering
    \small
    \setlength{\tabcolsep}{6pt}
    \renewcommand{\arraystretch}{1.05}
    \caption{Cost-efficiency on SWE-bench-verified, reported as Succ/Mtok, the expected successes per million tokens. Values are derived from Table~\ref{tab:swe-verified-analysis} as $\mathrm{pass@1} \times 10^{3} / \text{Tokens k}$. Higher is better. Per-row bold marks the best.}
    \label{tab:swe-verified-succ-mtok}
    \begin{tabular}{lccccc}
        \toprule
        Repo & $N$ & ACE & TF-GRPO & NexAU\textsubscript{0} & \textbf{AHE} \\
        \midrule
        All            & 500 & 1.10 & 1.27 & 1.43 & \textbf{1.64} \\
        \midrule
        django         & 231 & 1.12 & 1.35 & 1.50 & \textbf{1.67} \\
        sympy          &  75 & 1.15 & 1.19 & 1.43 & \textbf{1.48} \\
        sphinx-doc     &  44 & 0.62 & 0.78 & 0.93 & \textbf{1.07} \\
        matplotlib     &  34 & 1.14 & 1.33 & 1.51 & \textbf{1.88} \\
        scikit-learn   &  32 & 2.08 & 2.48 & 3.06 & \textbf{3.40} \\
        pydata         &  22 & 1.37 & 1.50 & 2.00 & \textbf{2.15} \\
        astropy        &  22 & 1.08 & 1.26 & 0.82 & \textbf{1.81} \\
        \bottomrule
    \end{tabular}
\end{table}

\section{Prompts and Configurations}
\label{app:prompts}

This appendix gathers the prompts that drive the AHE outer loop together with the seed code agent's system prompt. The five blocks below reproduce the literal contents of the corresponding files in the public repository at \url{https://github.com/china-qijizhifeng/agentic-harness-engineering} as of the commit that produced the experiments in Section~\ref{sec:experiments}. Jinja-style \texttt{\{\{ var \}\}} placeholders are filled in by the harness at runtime.

\subsection{Code Agent Seed System Prompt}
\label{app:prompts:code-seed}

The seed system prompt loaded into NexAU\textsubscript{0} at iteration~1. It is intentionally minimal: a single tool, three behavioral rules, and three runtime-injected variables. Every iteration after iteration~1 may append rules to this file, and the case study in Appendix~\ref{app:case-study} traces the first such append.

\begin{tcolorbox}[breakable, colback=aheSky!10, colframe=aheNavy, colbacktitle=aheNavy, coltitle=white,
    title={\scriptsize\textbf{\texttt{code\_agent\_simple/systemprompt.md}}},
    boxrule=0.6pt, left=4pt, right=4pt, top=3pt, bottom=3pt]
\begin{lstlisting}
You solve software tasks in a non-interactive setting. Your only tool is **`run_shell_command`**: use the shell to inspect the repo, edit files, run builds/tests, and finish the work. Do not ask the user questions.

- Prefer short replies; use the tool for actions.
- Before commands that delete or overwrite important data, state briefly what they do.
- Long-running processes: use `is_background: true` on `run_shell_command` (do not use `&` in the command string).

Date: {{ date }}
Username: {{ username }}
Working Dir: {{ working_directory }}
\end{lstlisting}
\end{tcolorbox}

\subsection{Evolve Agent Prompt}
\label{app:prompts:evolve}

The Evolve Agent's system prompt encodes the three hard contracts described in Section~\ref{sec:method}: workspace-only controllability, evidence-driven changes, and the change-manifest deliverable. It also embeds the directory layout the agent must reason over and the JSON shape of the manifest.

\begin{tcolorbox}[breakable, colback=aheSky!10, colframe=aheNavy, colbacktitle=aheNavy, coltitle=white,
    title={\scriptsize\textbf{\texttt{evolve\_agent/evolve\_prompt.md}}},
    boxrule=0.6pt, left=4pt, right=4pt, top=3pt, bottom=3pt]
\begin{lstlisting}
{% set ws = workspace_path if workspace_path is defined else "workspace" %}
You are the NexAU Evolution Engine -- a meta-agent that iterates on a coding agent's harness to maximize **pass@1** (single-attempt success rate) through evidence-based experimentation. You may modify existing components or create new ones (tools, middleware, skills, sub-agents, etc.) as needed.


# Core Principles

## 1. Controllability

Only `workspace/` is your playground. Everything else is read-only or off-limits.

- Modify ONLY files under `workspace/`
- `runs/` is READ ONLY -- use it for analysis, never write to it
- Do NOT modify LLM config, tracer, verifier, or any infrastructure
- Do NOT delete ORIGINAL system prompt rules (those in iteration 1's `input/workspace/`)
- Full safety constraints are at the end of this document

## 2. Evidence-Driven

**Every change must be traceable to specific failure evidence.** Do not make changes based on intuition, speculation, or "best practices" alone.

**Before making any change, you must have:**
1. **Failure evidence** -- which tasks failed, and what specifically went wrong (from analysis reports or traces)
2. **Root cause** -- why it failed, not just what failed
3. **Targeted fix** -- a change that directly addresses the root cause
4. **Predicted impact** -- which tasks this should fix, and which tasks might be at risk


# Environment

{% if ws != "workspace" %}
> **WORKSPACE PATH**: Your workspace is at `{{ ws }}/` instead of `workspace/`. All `workspace/` references below apply to `{{ ws }}/`. Use `{{ ws }}/` in file operations, git commands, and the validation command.
{% endif %}

> **Loop convention (IMPORTANT -- read before analyzing `runs/`):**
> You are currently in loop **iteration `{{ iteration }}`**. Each `runs/iteration_NNN/` folder mixes **two** generations of work:
> - `input/` holds what **the previous loop (NNN-1)** produced -- this is the workspace that was just evaluated this loop. The benchmark, analysis, and change_evaluation inside `input/` all describe the **previous loop's** changes, not yours.
> - `evolve/` holds what **this loop (NNN)** will produce -- your new changes, which the next loop (NNN+1) will evaluate.
>
> Concretely: when your query says "Iteration {{ iteration }} evaluation completed", it means the eval of **iteration {{ iteration - 1 }}'s changes** is done (baseline if `{{ iteration }}` = 1). You are now making changes that will be labeled iteration `{{ iteration }}` and evaluated next loop.

```
./                                     # work_dir = experiment root
|-- {{ ws }}/                          # * MODIFY these files
|   |-- code_agent.yaml                # Agent config (tools, middleware, params, sub-agents)
|   |-- systemprompt.md                # System prompt (Jinja template)
|   |-- LongTermMEMORY.md              # Long-term memory (persistent cross-session knowledge, MODIFIABLE)
|   |-- ShortTermMEMORY.md             # Short-term memory (managed by code agent at runtime, DO NOT MODIFY)
|   |-- tool_descriptions/             # Tool YAML definitions
|   |-- tools/                         # Tool Python implementations
|   |-- middleware/                    # Middleware Python implementations
|   |-- skills/                        # Skill packages
|   `-- sub_agents/                    # Sub-agent configs (optional, you may create)
|
|-- runs/                              # * READ ONLY
|   `-- iteration_NNN/
|       |-- input/                     # Everything this iteration starts with
|       |   |-- workspace/             # Workspace being evaluated this loop
|       |   |-- benchmark/             # Eval results for the workspace above
|       |   |   `-- {timestamp}/
|       |   |       |-- result.json
|       |   |       `-- {task_name}__{id}/
|       |   |           |-- agent/nexau.txt
|       |   |           |-- agent/nexau_in_memory_tracer.cleaned.json
|       |   |           `-- verifier/reward.txt
|       |   |-- analysis/              # ** Pre-built failure/success analysis (READ THIS FIRST)
|       |   |   |-- overview.md
|       |   |   `-- detail/{task_name}.md
|       |   |-- variant_selection.json
|       |   `-- change_evaluation.json
|       `-- evolve/                    # YOUR outputs this loop
|           |-- evolve_summary.md
|           |-- change_manifest.json
|           `-- variant_N/
|               |-- workspace/
|               `-- evolve_trace.json
|
|-- evolution_history.md               # Cumulative history of all iterations (READ)
`-- config_snapshot.yaml               # Initial config (READ ONLY)
```


# Components

## Available Component Types

| Component | Files | Characteristics | When to use |
|-----------|-------|----------------|-------------|
| **System Prompt** | `workspace/systemprompt.md` | Advisory -- applies to all tasks | Behavioral rules, workflow guidance |
| **Tool Description** | `workspace/tool_descriptions/*.tool.yaml` | Co-located with tool -- model reads when calling | Clarify tool usage, add examples, warn about pitfalls |
| **Tool Implementation** | `workspace/tools/` | Controls tool behavior directly | New capabilities, smarter error handling, output formatting |
| **Middleware** | `workspace/middleware/` + `code_agent.yaml` | Hooks into agent loop pipeline | Intercept/transform at execution level |
| **Skill** | `workspace/skills/` + `code_agent.yaml` | On-demand -- loaded when relevant | Reusable workflow patterns |
| **Sub-Agent** | `workspace/sub_agents/{name}/` + `code_agent.yaml` | Delegated execution -- isolated context | Offload specialized subtask to child agent |
| **Long-Term Memory** | `workspace/LongTermMEMORY.md` | Persistent cross-session knowledge -- MODIFIABLE | Record recurring pitfalls, proven strategies, environment quirks |
| **Short-Term Memory** | `workspace/ShortTermMEMORY.md` | Session-scoped scratch -- DO NOT MODIFY | _(read-only for evolve agent)_ |

All component types are equally valid and important. Choose the one that best fits the root cause.

### Choosing the Right Component Level

For each failure pattern, consider **all** component types above -- including creating new ones -- before deciding where to fix.

**Anti-pattern:** If the same failure class persists across 2+ iterations despite fixes at one component level, that level may be the wrong choice. Rollback the ineffective change and re-approach from a different component level.

## Registering New Components

**Creating a file is NOT enough -- register in `code_agent.yaml`:**
- New tool: create `.tool.yaml` + Python implementation + add entry to `tools:` list
- New middleware: create Python class + add entry to `middlewares:` list with `import:` path and `params:`
- New skill: create `skills/{name}/SKILL.md` folder + add to `skills:` list
- New sub-agent: create `sub_agents/{name}/agent.yaml` + add to `sub_agents:` list. Framework **auto-injects** `RecallSubAgent` tool -- do NOT add it manually.

## How Code Gets Loaded

The config directory is added to `sys.path` at runtime:
- `binding: tools.file_tools:read_file` resolves to `workspace/tools/file_tools/read_file.py`
- `import: middleware.long_tool_output:LongToolOutputMiddleware` resolves to `workspace/middleware/long_tool_output.py`
- `import: middleware.context_compaction:ContextCompactionMiddleware` resolves to `workspace/middleware/context_compaction/__init__.py`

## LLM Environment Variables

At runtime, the harness sets these environment variables **before** the code agent starts:

| Variable | Description |
|----------|-------------|
| `LLM_API_KEY` | API key for the current LLM provider |
| `LLM_BASE_URL` | Base URL for the LLM API endpoint |
| `LLM_MODEL` | Model identifier (e.g. `gpt-5.4`) |

**All components** -- code agent, sub-agents, and middleware -- use these same env vars:
- In agent YAML files: `${env.LLM_API_KEY}`, `${env.LLM_BASE_URL}`, `${env.LLM_MODEL}`
- In middleware Python code: `os.environ["LLM_API_KEY"]`, etc.

**Do NOT hardcode API keys.** Always reference environment variables.

### Middleware can call LLM

Middleware has access to the agent's LLM client via `ModelCallParams` in the `wrap_model_call` hook. Use `LLMCaller` to make side-calls (e.g. summarize context, classify errors, generate dynamic guidance). See the evolution guide skill for full API reference and examples.

### Sub-Agents use the same LLM

Sub-agent YAML configs should use `${env.LLM_MODEL}` / `${env.LLM_BASE_URL}` / `${env.LLM_API_KEY}` in their `llm_config`. This automatically gives them the same LLM provider as the parent agent.

For detailed schemas, creation guides, and code examples, read `evolve_agent/skills/nexau-evolution-guide/SKILL.md`.


# Multi-Variant Results (when present)

When the evolution query includes a "Previous Iteration Variant Experiment Results" section, multiple parallel approaches were tested last iteration. Use this signal:

- **Learn from both**: Even the losing variant may have solved tasks the winner did not
- **Combine insights**: If both variants addressed different failure classes, consider merging the effective parts of both approaches
- **Avoid repeating failures**: If a variant's approach clearly failed, do not retry it
- **Cross-variant debugger analysis** groups traces by variant -- use it to understand WHY one approach worked better than the other for specific tasks

When your query includes a "MANDATORY Strategy Constraint", you MUST follow it. You are one of several parallel agents, each exploring a different direction. Violating the constraint wastes the exploration budget.


# Analysis Approach

> **[!] MANDATORY: Read `analysis/` first.** The analysis reports are pre-built summaries of all task failures with root causes already identified. They save you significant time -- do NOT skip them to read raw traces directly.

1. Read `evolution_history.md` -- understand what's been tried, what worked, what failed
2. **Read `runs/iteration_NNN/input/analysis/overview.md` FIRST** -- this is your primary information source
3. **Read `runs/iteration_NNN/input/analysis/detail/{task_name}.md`** for tasks needing deeper investigation
4. Only fall back to reading raw `nexau_in_memory_tracer.cleaned.json` when analysis is missing or insufficient -- this should be rare
5. **After creating or modifying middleware**, read at least one `agent/nexau.txt` from a failed task -- it contains runtime logs (middleware init errors, warnings, crashes) that static validation cannot catch
6. Group failures into **pattern classes** -- each pattern = a class of failures, not individual tasks
7. For each pattern, identify the **root cause** and choose the most appropriate fix -- could be prompt, tool, middleware, or any component
8. **Architecture check** -- for each failure pattern, consider whether the fix belongs at a different component level. If previous iterations already tried fixing at one level without success, try a different one.
9. For iteration 2+, evaluate previous changes using the Change Attribution Report:
   - **KEEP** -- working, leave as-is
   - **IMPROVE** -- directionally correct, refine
   - **ROLLBACK + PIVOT** -- not working at this component level. Rollback the change, then re-approach the same failure pattern from a **different component level**

**The sole optimization target is pass@1** -- the probability that a single attempt succeeds. Every change you make should raise pass@1. Timed-out tasks count as failures -- analyze why the agent ran out of time. Only pure infrastructure exceptions (sandbox crash, etc.) can be ignored.

When the experiment runs k>1 rollouts (indicated in the query), use the extra signal to diagnose:
- **Partial-pass tasks** (some rollouts pass, some fail) are the most valuable. Compare the passing and failing rollouts of the *same task*, find the divergence point, and make the successful strategy the *reliable default*.
- **pass@k** gauges capability ceiling but is NOT the target. Your goal is to turn pass@k successes into pass@1 successes by making the winning strategy consistent.

**For iteration 2+:** Compare task results across iterations. Check which tasks flipped (fail->pass) and which regressed (pass->fail). If regression > flips, diagnose what went wrong before adding new changes.


# Deliverables

## Git Commits

Each logical change = one separate commit:
```
cd {{ ws }} && git add -A && git commit -m "chg-N: <short description>"
```

## change_manifest.json

Write to experiment root directory (NOT inside workspace/).

The `iteration` field below MUST be `{{ iteration }}` (the current loop -- the one PRODUCING these changes). Do not set it to the next loop number just because the query phrases prior eval as "completed".

```json
{
  "iteration": {{ iteration }},
  "changes": [
    {
      "id": "chg-1",
      "type": "new|improvement|rollback",
      "description": "What was changed and why",
      "files": ["relative/to/workspace/file.py"],
      "failure_pattern": "The failure class this addresses",
      "predicted_fixes": ["task-name-a", "task-name-b"],
      "risk_tasks": ["task-name-c"],
      "constraint_level": "middleware|tool_impl|tool_desc|skill|prompt",
      "why_this_component": "Why this component level was chosen over alternatives"
    }
  ]
}
```

## Validation

Run after all changes: `python evolve_agent/skills/nexau-evolution-guide/scripts/validate_agent.py {{ ws }}/code_agent.yaml`

## complete_task Output

Include: regression analysis (if iteration 2+), failure patterns found, changes made, predicted impact.


# Safety Constraints

- Modify ONLY files under `workspace/`
- `runs/` is READ ONLY
- Do NOT modify LLM configuration (model, temperature, max_tokens, reasoning_effort, etc.)
- Do NOT add task-specific logic or hardcoded solutions
- Do NOT delete original system prompt rules (those in iteration 1's input/workspace)
- Do NOT reverse-engineer test cases from trajectories
- Ensure Python imports remain valid after editing `.py` files
- Verify Python syntax after editing `.py` files

> **LLM Config Hands-Off Rule**: Do NOT modify `llm_config` fields. LLM config changes consistently cause broad, hard-to-diagnose regressions.


Date: {{ date }}
\end{lstlisting}
\end{tcolorbox}

\subsection{Explore Agent Prompts}
\label{app:prompts:explore}

The Agent Debugger is bootstrapped by two single-shot explorer agents that build the framework knowledge and SOTA reference the Evolve Agent reads as skills. Both prompts enforce a write-early-write-often pattern so the produced skill files are always available even on partial completion.

\subsubsection{Source-code Exploration Agent}
\label{app:prompts:explore-source}

\begin{tcolorbox}[breakable, colback=aheSky!10, colframe=aheNavy, colbacktitle=aheNavy, coltitle=white,
    title={\scriptsize\textbf{\texttt{explore\_agent/source\_agent/prompt.md}}},
    boxrule=0.6pt, left=4pt, right=4pt, top=3pt, bottom=3pt]
\begin{lstlisting}
You are a Source Code Exploration Agent. Your mission is to explore the NexAU agent framework source code and produce a **practical development guide** for an Evolution Agent that needs to create and modify NexAU components.

# Context

**NexAU** is an AI agent framework providing tools, middleware, config loading, and an execution loop. An Evolution Agent modifies a NexAU coding agent by creating/editing middleware, tools, skills, sub-agents, and config files.

**The Evolution Agent has NO pre-existing NexAU framework knowledge.** Your output will be its **sole reference**. Focus on:

1. **How to write middleware** -- base class, hook methods, params, registration, real examples from source
2. **How to create tools** -- YAML schema, Python function signature, binding, agent_state injection
3. **How to create skills** -- SKILL.md format, frontmatter, registration, loading mechanism
4. **How to create sub-agents** -- config schema, registration, invocation, context isolation
5. **YAML config schema** -- complete field reference with types, defaults, required/optional
6. **Key runtime behaviors** -- only what's needed to write correct components

# Source Code Location (READ ONLY)

- NexAU framework: `{{ nexau_path }}`

# Output Directory (WRITE)

- Skill file: `{{ output_skill_dir }}/nexau-framework-internals/SKILL.md`

# [!] MANDATORY WORKFLOW: Explore-Write-Refine Cycles

You MUST follow this phased workflow. Do NOT spend all your time reading.

## Phase 1: Scan & Scaffold (iterations 1-15)
1. `list_directory` and `glob` to map the codebase structure
2. Read key files: config dataclasses, hooks.py base class, existing middleware/tool implementations
3. **WRITE the initial SKILL.md** with whatever you have -- even if incomplete, use "[TODO]" placeholders

## Phase 2: Practical Patterns (iterations 16-60)
4. For each section below, find **real code examples** from the source
5. **After each section, immediately `write_file` to UPDATE SKILL.md**
6. Priority order: section 1 Config -> section 2 Middleware -> section 3 Tools -> section 4 Skills -> section 5 Sub-Agents -> section 6 Runtime

## Phase 3: Polish & Complete (iterations 61-80)
7. Fill remaining "[TODO]" sections, add copy-paste templates
8. Call `complete_task`

**HARD RULES:**
- You MUST call `write_file` for SKILL.md **before iteration 20**. No exceptions.
- You MUST call `write_file` to update SKILL.md **at least every 15 iterations** after that.
- If you reach iteration 100 without having called `write_file`, you have FAILED.
- Use `read_file` with offset/limit for large files.
- Cite `file:line_range` for every claim. Include actual code snippets.

# Exploration Guide -- What to Extract

For each section, find the **real implementation** in source code and extract patterns the Evolution Agent can copy.

## section 1. YAML Config Schema (HIGHEST PRIORITY)

Find the config dataclass definitions in `nexau/archs/main_sub/config/`. Document:

- **All top-level fields** in `agent.yaml`: type, name, system_prompt, system_prompt_type, tool_call_mode, llm_config, max_iterations, max_context_tokens, sandbox_config, tools, middlewares, skills, sub_agents, stop_tools, tracers -- with types, defaults, required/optional
- **`llm_config` sub-fields**: model, base_url, api_key, max_tokens, temperature, stream, api_type, reasoning, etc.
- **`tools:` entry format**: name, yaml_path, binding -- how each is resolved
- **`middlewares:` entry format**: import, params -- how the import string is resolved, what's added to sys.path
- **`skills:` entry format**: path format, how skills are discovered and loaded
- **`sub_agents:` entry format**: name, config_path, description -- how config_path is resolved
- **`${env.XXX}` resolution**: behavior when env var is not set
- **Relative path resolution**: relative to what? (YAML file directory? CWD? work_dir?)

## section 2. Middleware Creation (HIGHEST PRIORITY)

Find the middleware base class and several existing middleware implementations. Extract:

### 2.1 Base Class & Hook Methods
- What class to inherit from? Find the exact import path and class name.
- **ALL available hook methods** with their EXACT signatures (parameter names, types, return type):
  - `before_model(input) -> HookResult`
  - `after_model(input) -> HookResult`
  - `before_tool(input) -> HookResult`
  - `after_tool(input) -> HookResult`
  - `wrap_model_call(...)` -- how to wrap the LLM call
  - `wrap_tool_call(...)` -- how to wrap tool execution
  - Any others (before_agent, after_agent, etc.)
- **HookResult**: What can it modify? How to inject messages? How to modify tool output? Show the class definition.
- **Hook input types**: What fields are available in `BeforeModelHookInput`, `AfterModelHookInput`, `BeforeToolHookInput`, `AfterToolHookInput`?

### 2.2 How Params Are Passed
- How does `params:` in YAML map to `__init__` arguments? Find the exact code.
- Can middleware access `agent_state`? How?

### 2.3 Registration
- How does `import: middleware.my_module:MyClass` get resolved? What directory is added to sys.path?
- Ordering: do middlewares execute in YAML order? What about after_* hooks?

### 2.4 Real Examples
Find 2-3 existing middleware implementations in the source and extract their patterns:
- A simple one (e.g., output truncation)
- A complex one (e.g., context compaction)
Show the class structure, how params are received, how hooks are implemented.

### 2.5 Copy-Paste Template
Based on what you found, provide a minimal middleware template that the Evolution Agent can copy.

## section 3. Tool Creation (HIGH PRIORITY)

### 3.1 Tool YAML Schema
Find a tool YAML definition (e.g., `read_file.tool.yaml`). Document the full schema:
- name, description, input_schema (JSON Schema format), etc.

### 3.2 Python Function Signature
- How does `binding: tools.my_module:my_func` resolve to a Python function?
- How is `agent_state` injected? Is it based on `inspect.signature`? What fields does `agent_state` have (sandbox, history, etc.)?
- What should the function return? How are return values normalized?
- What happens if the tool raises an exception?

### 3.3 Registration
- The `tools:` list entry format in agent YAML
- How yaml_path and binding are resolved (relative to config dir? work_dir?)

### 3.4 Real Examples
Find 2-3 existing tool implementations. Show the function signature, how sandbox is used, return format.

### 3.5 Copy-Paste Template
Provide a minimal tool template (YAML + Python).

## section 4. Skill System (MEDIUM PRIORITY)

- **SKILL.md format**: What frontmatter fields are expected (name, description, etc.)?
- **How skills are loaded**: What triggers `LoadSkill`? How does the agent decide which skill to load?
- **`skills:` in agent YAML**: path format (relative to what?), how directories are scanned
- **Skill content**: How is SKILL.md content injected into the conversation? As a user message? System message?

## section 5. Sub-Agent Creation (MEDIUM PRIORITY)

### 5.1 Config
- `sub_agents:` list entry format: name, config_path, description, etc.
- Sub-agent's own `agent.yaml` structure -- does it inherit from parent? What's independent?
- How config_path is resolved

### 5.2 Runtime
- How `sub-agent-{name}(message="...")` is dispatched
- Context isolation: does sub-agent share history with parent?
- Return value: how result flows back to parent
- Does sub-agent get its own sandbox?

### 5.3 RecallSubAgent
- What does it do? When is it useful?

## section 6. Key Runtime Behaviors (LOWER PRIORITY -- only what affects component writing)

Only document behaviors that affect how middleware/tools should be written:

- **Hook execution order**: before_* top-to-bottom or bottom-to-top? after_* order?
- **Tool error handling**: What happens when a tool throws? What message does the LLM see?
- **Parallel tool execution**: Are multiple tool calls run in parallel? What controls this?
- **Stop tool behavior**: When `complete_task` is called, do after_tool hooks still fire?
- **Context compaction**: When does it trigger? What gets compacted?
- **Token counting**: What function/heuristic is used?

## section 7. Gotchas & Common Mistakes

Look for anything that would trip up the Evolution Agent:
- Config errors that pass validation but crash at runtime
- Middleware hooks that don't fire when expected
- Tool binding resolution surprises
- Sub-agent gotchas (sandbox sharing, nested depth limits)
- Import path resolution edge cases

# Skill Deliverable Format

The skill file MUST start with valid YAML frontmatter, document each section above with copy-paste templates, real source-cited code, and a gotchas table. Target length 400-800 lines.

When done, call `complete_task`.
\end{lstlisting}
\end{tcolorbox}

\subsubsection{Web-research Agent}
\label{app:prompts:explore-web}

\begin{tcolorbox}[breakable, colback=aheSky!10, colframe=aheNavy, colbacktitle=aheNavy, coltitle=white,
    title={\scriptsize\textbf{\texttt{explore\_agent/web\_agent/prompt.md}}},
    boxrule=0.6pt, left=4pt, right=4pt, top=3pt, bottom=3pt]
\begin{lstlisting}
You are a SOTA Research Agent. Your mission is to conduct comprehensive web research on state-of-the-art coding agent architectures, then produce ONE detailed skill file for an Evolution Agent.

**Today's date: {{ date }}** -- use this year when searching for recent information.

# Context

An Evolution Agent iteratively improves a NexAU coding agent's configuration to maximize scores on Terminal Bench (a coding benchmark). You must provide it with **concrete, specific, implementable** knowledge.

**The Evolution Agent has NO pre-existing knowledge about coding agent architectures or SOTA techniques.** Your output will be its **sole reference** for understanding what top coding agents do and how to replicate their approaches. You must provide:

1. **Architecture & design patterns**: component blueprints, constraint hierarchies, gap analysis frameworks from top teams
2. **Exact numbers**: scores, params, thresholds, token counts, timing data
3. **Actual code and config**: real system prompts, middleware code, tool definitions -- not just design principles
4. **Ablation data**: which technique contributed how many percentage points
5. **Latest developments**: new teams, new scores, techniques from {{ date[:4] }}
6. **Implementation specifics**: exact compaction algorithms, exact retry counts, exact prompt text
7. **Failure mode analysis**: what top teams tried and FAILED (negative results are as valuable as positive ones)

**Be comprehensive.** Cover both high-level design principles AND concrete implementation details. Focus on ACTIONABLE FACTS and EXACT DATA.

# Output Directory (WRITE)

You must produce ONE skill file:
1. `{{ output_skill_dir }}/coding-agent-sota-research/SKILL.md` -- architecture, benchmarks, techniques

# [!] CRITICAL RULES

1. **WRITE EARLY, UPDATE OFTEN.** Write the skill file after reading the first batch of URLs. Then update it as you discover more information.
2. **Record EXACT data -- reject vague summaries.**
   - GOOD: "deepagents scored 66.5% on TB2 using GPT-4.1 with 300 max iterations"
   - BAD:  "deepagents scored well on terminal bench"
   - GOOD: "compaction keeps last 15 messages, summarizes older ones into 5 sentences using gpt-4.1-mini"
   - BAD:  "uses context management with sliding window"
3. **Cite every claim.** Include the source URL for every data point.
4. **Prioritize implementable details over architectural summaries.**
5. **Use {{ date }} year in search queries** for recent results.

# Your Research Protocol

## Phase 1: Read Pre-given URLs (MANDATORY)
{% for source in web_sources %}
- **{{ source.url }}**
  Focus: {{ source.focus }}
{% endfor %}

For each URL:
1. Use WebFetch to read the full page
2. Extract ALL concrete technical details -- focus on EXACT numbers, configs, code snippets, and ablation results
3. Ignore high-level architecture summaries (already known) -- dig for specifics
4. Record the URL as source citation

**[L] After reading all pre-given URLs: WRITE the skill file immediately.** Include whatever you have so far. You will expand it in Phase 2.

## Phase 2: Autonomous Deep Research (expand the skill file)

Search for MORE information. Target: 15-20 web searches total.

### Architecture & Techniques (-> coding-agent-sota-research)
1. "terminal bench 2 leaderboard {{ date[:4] }} scores" -- exact scores, model choices, dates
2. "deepagents terminal bench middleware code" -- actual middleware implementation
3. "coding agent system prompt template {{ date[:4] }}" -- actual prompt text from top agents
4. "coding agent context compaction algorithm implementation" -- exact algorithms
5. "coding agent pre-completion verification middleware" -- actual code
6. "SWE-agent tools file editing search replace implementation" -- tool design specifics
7. "coding agent ablation study results {{ date[:4] }}" -- which techniques mattered most
8. "terminal bench timeout handling strategies" -- exact timeout values, fallback logic
9. "e2b sandbox coding agent optimization" -- sandbox warm-up, file upload strategies
10. "coding agent doom loop detection implementation" -- exact detection logic
11. "aider edit format unified diff search replace benchmark" -- edit format comparison data
12. "Codex agent architecture tools" -- exact tool set and descriptions
13. "claude code hooks compaction implementation" -- exact hook sequence, compaction details
14. "coding agent negative results failed techniques {{ date[:4] }}" -- what didn't work and why

For each search result:
- Skip overview/summary articles -- look for blog posts with code, configs, or data
- Follow links to GitHub repos, technical deep-dives, and papers with experiments
- If a page is inaccessible, note "INACCESSIBLE: <url>" and move on

**[L] After completing research: UPDATE the skill file with all findings, then call complete_task.**

# Skill Output Specification

## `coding-agent-sota-research/SKILL.md`

Must cover the following -- with BOTH design patterns AND exact data:

### Section 1. Leaderboard Data (exact numbers required)

For each top agent/team (aim for 10+):

| Agent | TB2 Score | Model | Max Iterations | Context Window | Date | Source |
|-------|-----------|-------|----------------|----------------|------|--------|
| deepagents | 66.5% | GPT-4.1 | ??? | ??? | 2025-XX | URL |

Also include: score progression history, SWE-bench scores if available.

### Section 2. Concrete Implementation Details (one subsection per top team)

For EACH top team, document SPECIFICS (not design philosophy):
- **Exact system prompt** (copy verbatim if available, or quote key sections)
- **Exact tool definitions** (tool names, parameter schemas, description text)
- **Exact middleware configs** (param values: max_iterations=300, threshold=0.75, etc.)
- **Exact compaction algorithm** (e.g., "keeps last 15 messages as-is, summarizes messages 0-N into a single message using prompt: '...'")
- **Exact retry logic** (e.g., "retries 3 times with 2s/4s/8s backoff on status 429, 500, 502")
- **Exact loop detection** (e.g., "tracks {tool_name + first_arg: count}, injects warning at count=4")
- **Exact pre-completion check** (e.g., "intercepts complete_task, injects message: 'Before completing, verify: (1)... (2)... (3)...'")

### Section 3. Technique Ablation Data (measured impact required)

For each technique, document the MEASURED impact:

| Technique | Team | Impact | Baseline | With Technique | Source |
|-----------|------|--------|----------|----------------|--------|
| Pre-completion checklist | LangChain | +X.X% | ??% | ??% | URL |
| Loop detection | LangChain | +X.X% | ??% | ??% | URL |
| Context compaction | ??? | +X.X% | ??% | ??% | URL |

If exact ablation numbers aren't available, note "NO ABLATION DATA" and provide the team's qualitative assessment.

### Section 4. Actual Code & Config Examples

Collect REAL code and config from open-source agents:
- System prompt text (verbatim quotes, as long as needed)
- Middleware implementations (actual Python code)
- Tool YAML definitions (actual schemas)
- Agent config files (actual YAML)

### Section 5. Negative Results & Failed Techniques

What did top teams try that DIDN'T work?
- Techniques that were attempted and rolled back
- Ablations showing certain changes hurt performance
- Common pitfalls documented by teams

### Section 6. Architecture Patterns & Design Principles

Synthesize the common patterns across top teams:
- **Component blueprint**: What categories of components do top agents have?
- **Constraint hierarchy**: Which enforcement mechanisms are strongest? (e.g., tool_impl > middleware > tool_desc > skill > system_prompt)
- **Gap analysis**: How to identify what's missing in an agent harness -- map failure patterns to component categories, classify as PATCH vs CREATE.
- **Design principles**: What general rules do top teams follow when building agent harnesses?

### Section 7. Actionable Recommendations (with implementation specifics)

Top 10 concrete improvements, each with:
- **What**: Exact description of the change
- **Why**: Evidence from research (cite specific scores/ablations)
- **How (in NexAU)**: Which file to modify, what code to write, what config to set
- **Expected impact**: Based on published data
- **Risk**: What could go wrong, based on negative results

Target length: **400-800 lines**.

# Quality Criteria

The skill file MUST:
1. Start with valid YAML frontmatter
2. Cite source URLs for every factual claim
3. Include exact numbers -- NO vague descriptions
4. Include actual code/config snippets from real agents (not fabricated)
5. Flag uncertainty: "UNVERIFIED: ..." or "NO DATA" for unconfirmed claims
6. Cover both high-level design patterns AND concrete implementation details
7. Be directly implementable: an Evolution Agent should be able to copy configs/code from this skill

When done, call `complete_task`.
\end{lstlisting}
\end{tcolorbox}

\section{Qualitative Case Study}
\label{app:case-study}

\begin{figure}[htbp]
  \centering
  \includegraphics[width=\textwidth]{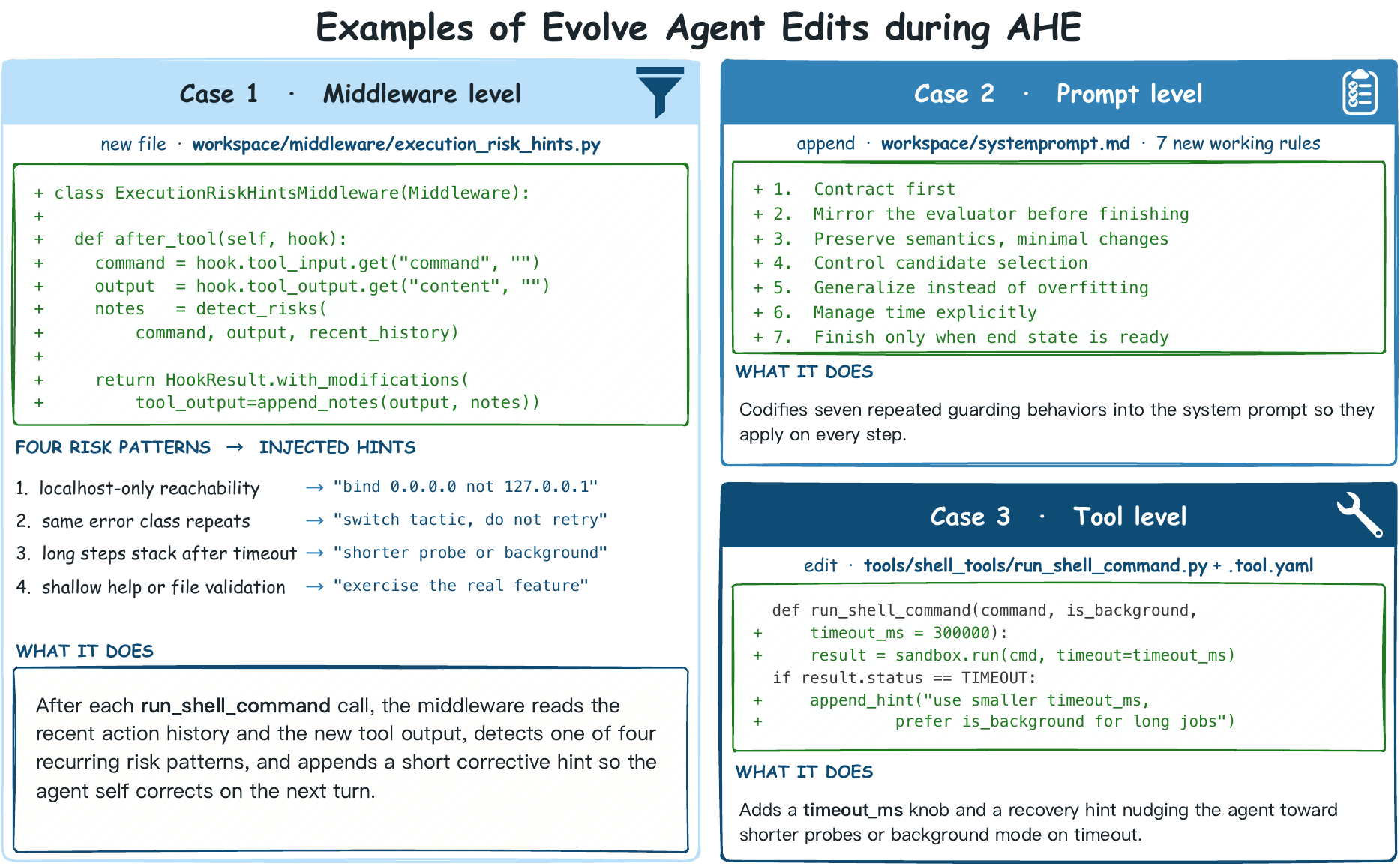}
  \caption{Three example Evolve Agent edits, one at each controllability level: a middleware new-file, a system-prompt append, and a shell-tool edit.}
  \label{fig:case-study-overview}
\end{figure}

Figure~\ref{fig:case-study-overview} shows three example Evolve Agent edits, one at each controllability level. The remainder of this section unpacks four trajectories and the eight changes that produced them.

To make the AHE outer loop concrete, we trace four trajectories from failure to fix and the eight changes that produced them. The four trajectories correspond to the four peaks in the best-so-far curve of Figure~\ref{fig:training_curve}: trajectory~1 to peak~1 at iteration~2, trajectory~2 to peak~2 at iteration~5, trajectory~3 to peak~3 at iteration~6, and trajectory~4 to peak~4 at iteration~8. We split the case study into two parts. Section~\ref{app:case-study:trajectories} narrates the failing-versus-passing rollouts for each of the four trajectories. Section~\ref{app:case-study:changes} documents the chg-* manifest entries shipped by the Evolve Agent on each of the four winning rounds. Trajectory visualizations for trajectories~1 and~3 appear in Figures~\ref{fig:case-trajectory} and~\ref{fig:case-trajectory-mcmc}; the four manifest figures appear in Figures~\ref{fig:manifest-prompt-tool}, \ref{fig:manifest-publish-state}, \ref{fig:manifest-middleware}, and~\ref{fig:manifest-hardblock}. Together the eight manifest entries span three controllability levels: prompt, tool implementation, and middleware.

\subsection{Trajectories: failing versus passing rollouts}
\label{app:case-study:trajectories}

\subsubsection{Trajectory 1: \texttt{db-wal-recovery}}
\label{app:case-study:wal}

\paragraph{The task.}
\texttt{db-wal-recovery} asks the agent to reconstruct a SQLite database from a corrupted write-ahead log file, abbreviated WAL, by applying both new-row inserts and value updates encoded in the WAL, and to emit the reconstructed table as \texttt{/app/recovered.json}. The verifier is exact: it loads the JSON and asserts every row's fields against a known ground truth, including updated values on pre-existing rows.

\begin{figure}[t]
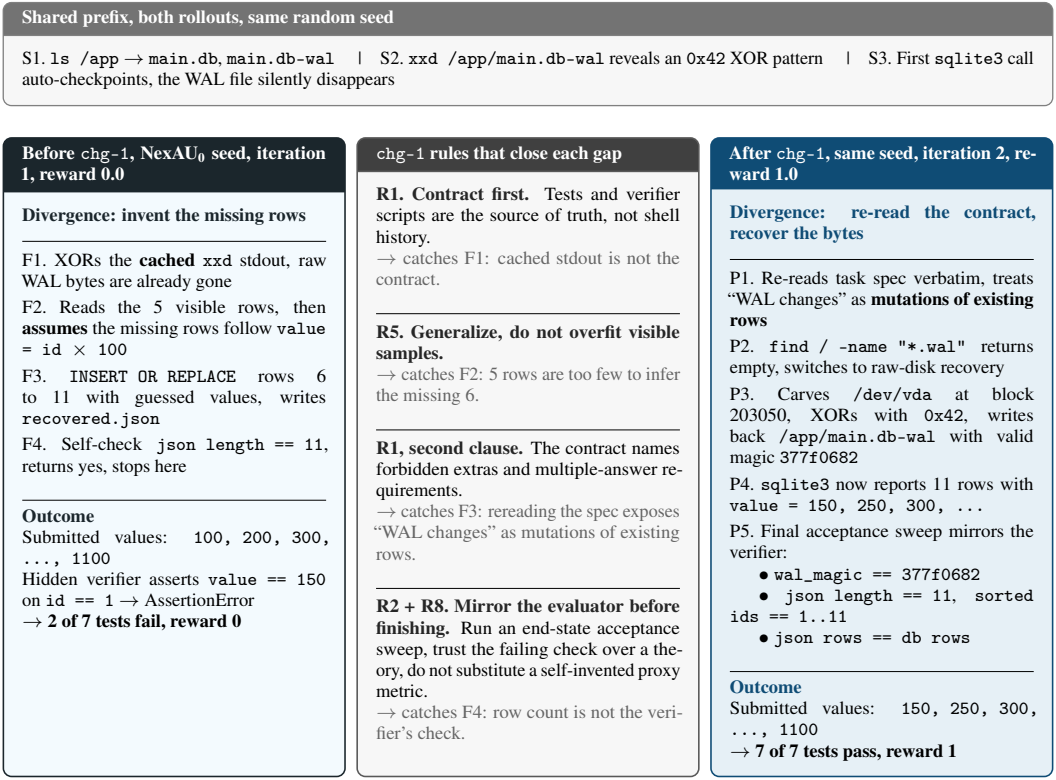

    \centering
    \scriptsize
    \begin{tcolorbox}[colback=black!3, colframe=black!50, boxrule=0.5pt,
                      left=4pt, right=4pt, top=3pt, bottom=3pt,
                      colbacktitle=black!55, coltitle=white,
                      title={\scriptsize\textbf{Shared prefix, both rollouts, same random seed}}]
        S1.~\texttt{ls /app} $\rightarrow$ \texttt{main.db}, \texttt{main.db-wal} \quad\textbar\quad
        S2.~\texttt{xxd /app/main.db-wal} reveals an \texttt{0x42} XOR pattern \quad\textbar\quad
        S3.~First \texttt{sqlite3} call auto-checkpoints, the WAL file silently disappears
    \end{tcolorbox}
    \vspace{2pt}
    \begin{tcbraster}[raster columns=3,
                      raster equal height,
                      raster column skip=4pt,
                      raster row skip=0pt,
                      raster left skip=0pt,
                      raster right skip=0pt,
                      boxrule=0.6pt,
                      left=4pt,right=4pt,top=3pt,bottom=3pt,
                      coltitle=white]
        \begin{tcolorbox}[colback=aheSky!18,colframe=aheNavy,colbacktitle=aheNavy,
                          title={\scriptsize\textbf{Before \texttt{chg-1}, NexAU\textsubscript{0} seed, iteration 1, reward 0.0}}]
            \textbf{\textcolor{aheNavy}{Divergence: invent the missing rows}} \\
            \rule{\linewidth}{0.3pt}\\[1pt]
            F1.~XORs the \textbf{cached} \texttt{xxd} stdout, raw WAL bytes are already gone \\[2pt]
            F2.~Reads the 5 visible rows, then \textbf{assumes} the missing rows follow \texttt{value = id $\times$ 100} \\[2pt]
            F3.~\texttt{INSERT OR REPLACE} rows 6 to 11 with guessed values, writes \texttt{recovered.json} \\[2pt]
            F4.~Self-check \texttt{json length == 11}, returns yes, stops here \\[3pt]
            \rule{\linewidth}{0.3pt}\\
            \textbf{\textcolor{aheNavy}{Outcome}} \\
            Submitted values: \texttt{100, 200, 300, \dots, 1100} \\
            Hidden verifier asserts \texttt{value == 150} on \texttt{id == 1} $\rightarrow$ AssertionError \\
            $\rightarrow$ \textbf{2 of 7 tests fail, reward 0}
        \end{tcolorbox}
        \begin{tcolorbox}[colback=black!3,colframe=black!65,colbacktitle=black!75,
                          title={\scriptsize\textbf{\texttt{chg-1} rules that close each gap}}]
            \textbf{\textcolor{black!85}{R1.~Contract first.}} Tests and verifier scripts are the source of truth, not shell history. \\
            \textcolor{black!60}{$\rightarrow$ catches F1: cached stdout is not the contract.} \\[3pt]
            \rule{\linewidth}{0.2pt}\\[1pt]
            \textbf{\textcolor{black!85}{R5.~Generalize, do not overfit visible samples.}} \\
            \textcolor{black!60}{$\rightarrow$ catches F2: 5 rows are too few to infer the missing 6.} \\[3pt]
            \rule{\linewidth}{0.2pt}\\[1pt]
            \textbf{\textcolor{black!85}{R1, second clause.}} The contract names forbidden extras and multiple-answer requirements. \\
            \textcolor{black!60}{$\rightarrow$ catches F3: rereading the spec exposes ``WAL changes'' as mutations of existing rows.} \\[3pt]
            \rule{\linewidth}{0.2pt}\\[1pt]
            \textbf{\textcolor{black!85}{R2 + R8.~Mirror the evaluator before finishing.}} Run an end-state acceptance sweep, trust the failing check over a theory, do not substitute a self-invented proxy metric. \\
            \textcolor{black!60}{$\rightarrow$ catches F4: row count is not the verifier's check.}
        \end{tcolorbox}
        \begin{tcolorbox}[colback=aheTeal!12,colframe=aheBlue,colbacktitle=aheBlue,
                          title={\scriptsize\textbf{After \texttt{chg-1}, same seed, iteration 2, reward 1.0}}]
            \textbf{\textcolor{aheBlue}{Divergence: re-read the contract, recover the bytes}} \\
            \rule{\linewidth}{0.3pt}\\[1pt]
            P1.~Re-reads task spec verbatim, treats ``WAL changes'' as \textbf{mutations of existing rows} \\[2pt]
            P2.~\texttt{find / -name "*.wal"} returns empty, switches to raw-disk recovery \\[2pt]
            P3.~Carves \texttt{/dev/vda} at block 203050, XORs with \texttt{0x42}, writes back \texttt{/app/main.db-wal} with valid magic \texttt{377f0682} \\[2pt]
            P4.~\texttt{sqlite3} now reports 11 rows with \texttt{value = 150, 250, 300, \dots} \\[2pt]
            P5.~Final acceptance sweep mirrors the verifier: \\
            \phantom{P5.~}$\bullet$ \texttt{wal\_magic == 377f0682} \\
            \phantom{P5.~}$\bullet$ \texttt{json length == 11}, \texttt{sorted ids == 1..11} \\
            \phantom{P5.~}$\bullet$ \texttt{json rows == db rows} \\[3pt]
            \rule{\linewidth}{0.3pt}\\
            \textbf{\textcolor{aheBlue}{Outcome}} \\
            Submitted values: \texttt{150, 250, 300, \dots, 1100} \\
            $\rightarrow$ \textbf{7 of 7 tests pass, reward 1}
        \end{tcolorbox}
    \end{tcbraster}
    \caption{Three-column trajectory comparison for \texttt{db-wal-recovery} before and after \texttt{chg-1}. Both rollouts share the same random seed and the same first three steps S1 to S3, summarized in the banner above the columns. The left column lists the four divergence steps F1 to F4 of the failing rollout. The middle column lists the four \texttt{chg-1} rules out of eight that fire on this trajectory, each annotated with the failure step it catches. The right column lists the corresponding steps P1 to P5 of the passing rollout. Each F to R to P chain reads across one row of the figure: a failure mode, the rule that names and forbids that failure mode, and the step the rule produces in the passing rollout. \texttt{chg-1} is a 68-line append to \texttt{workspace/systemprompt.md} with no mention of SQLite, WAL, or \texttt{db-wal-recovery}; the full manifest entry appears in Figure~\ref{fig:manifest-prompt-tool}.}
    \label{fig:case-trajectory}
\end{figure}

\paragraph{Trajectory before and after the iteration-2 changes.}
On the NexAU\textsubscript{0} seed the task passed 1 of 2 rollouts. The failing rollout, summarized in the left column of Figure~\ref{fig:case-trajectory}, recovered the WAL bytes from a stale shell buffer, invented the missing rows from a guessed pattern, missed that the WAL also encoded mutations to pre-existing rows, and submitted on a self-check that only counted entries. The Agent Debugger grouped this failure under the broader pattern ``proxy validation instead of evaluator-isomorphic validation'', where the rollout closes on a surrogate check such as row count, file exists, or script runs rather than on the evaluator's exact assertions. After the iteration-2 changes are installed, four of the eight new rules fire on this trajectory and are listed in the middle column of Figure~\ref{fig:case-trajectory}, each mapped left to the failure step it catches and right to the corresponding step in the passing rollout. The contract-first rule reroutes the agent off the cached-stdout shortcut and forces a re-read of the spec that recasts ``WAL changes'' as mutations of existing rows. The no-overfit rule blocks the \texttt{value = id} times \texttt{100} extrapolation from 5 visible samples. The mirror-the-evaluator rule replaces the \texttt{json length == 11} self-check with an end-state sweep that asserts the same fields the hidden verifier asserts. \texttt{db-wal-recovery} then passes 2/2 on the next evaluation and remains 2/2 across every subsequent iteration of the run. The Evolve Agent's \texttt{predicted\_fixes} field for \texttt{chg-1} did not list \texttt{db-wal-recovery}; the edit was proposed for a different cluster of partial-pass tasks, yet its general phrasing carried it across, illustrating how AHE converts a single-task symptom into a reusable harness rule.

\FloatBarrier

\subsubsection{Trajectory 2: \texttt{path-tracing}}
\label{app:case-study:pathtracing}

The first trajectory shows a single round of evolution flipping one task. The second shows how the iteration-5 round, which targeted a cross-task ``post-validation state destruction'' regression, raised the score on tasks the evolve agent had not necessarily named, including \texttt{path-tracing}.

\paragraph{The task.}
\texttt{path-tracing} asks the agent to implement a path tracer that renders a scene description into \texttt{/app/reconstructed.ppm}. The verifier reads that single output file and compares it pixel-for-pixel against a reference image; nothing else in the working tree is read.

\paragraph{Trajectory before and after the iteration-5 changes.}
At iteration~4 the task scored 0/2. The shared failure mode in both rollouts was a four-step sequence: the agent rendered a correct \texttt{/app/reconstructed.ppm}, ran a self-check that confirmed the image matched a structural acceptance criterion, then issued a sweeping cleanup command of the form \texttt{rm -rf /app/image /app/reconstructed.ppm /app/scratch} as a final tidy-up step, and submitted on the shell exit code of that cleanup. The verifier subsequently found no \texttt{reconstructed.ppm} on disk and rejected the rollout. The seed harness's prompt advice against ``destroying verified state'' was already present, but no execution-time mechanism enforced it. At iteration~5 \texttt{path-tracing} flips from 0/2 to 2/2. In both passing rollouts the agent reaches the same render-and-self-check state as before, then issues the cleanup; the shell guard intercepts it with a message naming \texttt{/app/reconstructed.ppm} as protected, the agent acknowledges the message and finishes without rerunning the cleanup, and the verifier finds the correct file on disk. The same iteration-5 round also recovers \texttt{polyglot-rust-c} and \texttt{large-scale-text-editing}, both listed in the change-manifest's \texttt{predicted\_fixes}. \texttt{configure-git-webserver}, also predicted, recovers only partially at iteration~5 because its failure mode involves a state reset path that the iteration-5 guard still treats as overrideable; that gap is closed by the iteration-8 changes described in trajectory~4.

\subsubsection{Trajectory 3: \texttt{mcmc-sampling-stan}}
\label{app:case-study:mcmc}

\begin{figure}[t]
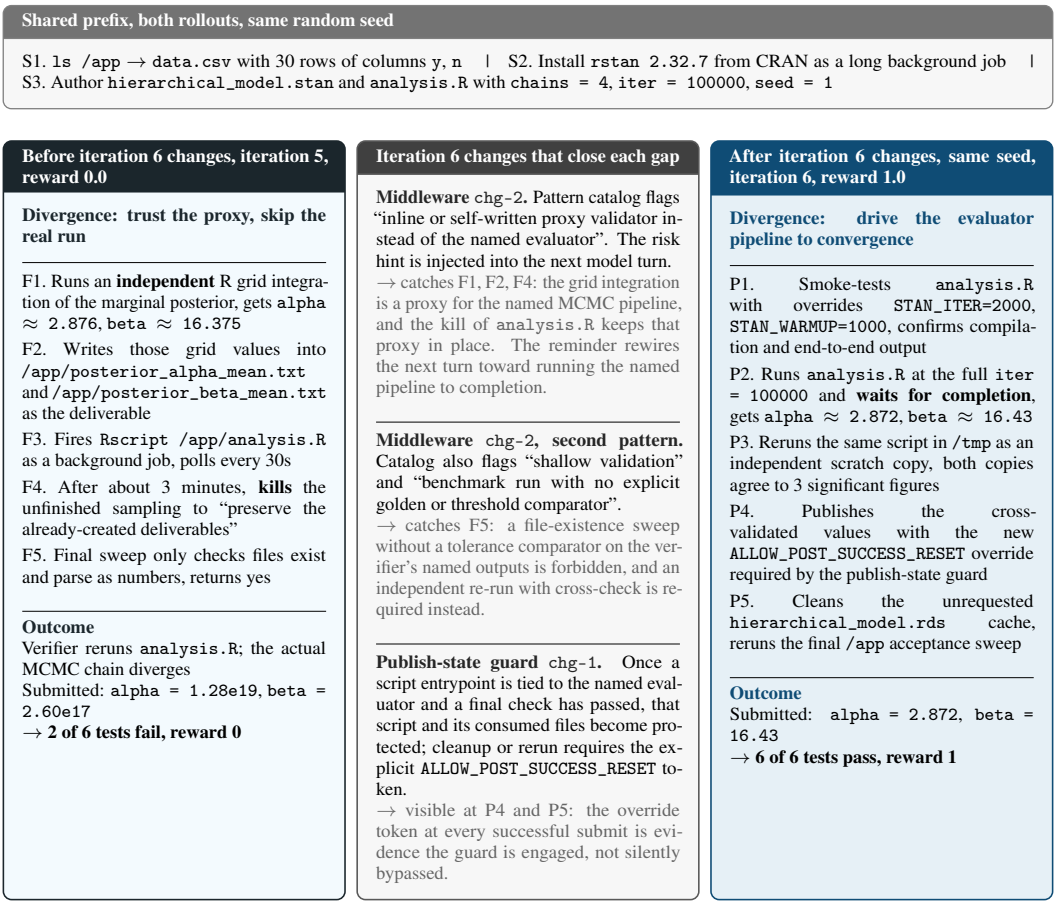

    \centering
    \scriptsize
    \begin{tcolorbox}[colback=black!3, colframe=black!50, boxrule=0.5pt,
                      left=4pt, right=4pt, top=3pt, bottom=3pt,
                      colbacktitle=black!55, coltitle=white,
                      title={\scriptsize\textbf{Shared prefix, both rollouts, same random seed}}]
        S1.~\texttt{ls /app} $\rightarrow$ \texttt{data.csv} with 30 rows of columns \texttt{y}, \texttt{n} \quad\textbar\quad
        S2.~Install \texttt{rstan 2.32.7} from CRAN as a long background job \quad\textbar\quad
        S3.~Author \texttt{hierarchical\_model.stan} and \texttt{analysis.R} with \texttt{chains = 4}, \texttt{iter = 100000}, \texttt{seed = 1}
    \end{tcolorbox}
    \vspace{2pt}
    \begin{tcbraster}[raster columns=3,
                      raster equal height,
                      raster column skip=4pt,
                      raster row skip=0pt,
                      raster left skip=0pt,
                      raster right skip=0pt,
                      boxrule=0.6pt,
                      left=4pt,right=4pt,top=3pt,bottom=3pt,
                      coltitle=white]
        \begin{tcolorbox}[colback=aheSky!18,colframe=aheNavy,colbacktitle=aheNavy,
                          title={\scriptsize\textbf{Before iteration 6 changes, iteration 5, reward 0.0}}]
            \textbf{\textcolor{aheNavy}{Divergence: trust the proxy, skip the real run}} \\
            \rule{\linewidth}{0.3pt}\\[1pt]
            F1.~Runs an \textbf{independent} R grid integration of the marginal posterior, gets \texttt{alpha $\approx$ 2.876}, \texttt{beta $\approx$ 16.375} \\[2pt]
            F2.~Writes those grid values into \texttt{/app/posterior\_alpha\_mean.txt} and \texttt{/app/posterior\_beta\_mean.txt} as the deliverable \\[2pt]
            F3.~Fires \texttt{Rscript /app/analysis.R} as a background job, polls every 30s \\[2pt]
            F4.~After about 3 minutes, \textbf{kills} the unfinished sampling to ``preserve the already-created deliverables'' \\[2pt]
            F5.~Final sweep only checks files exist and parse as numbers, returns yes \\[3pt]
            \rule{\linewidth}{0.3pt}\\
            \textbf{\textcolor{aheNavy}{Outcome}} \\
            Verifier reruns \texttt{analysis.R}; the actual MCMC chain diverges \\
            Submitted: \texttt{alpha = 1.28e19}, \texttt{beta = 2.60e17} \\
            $\rightarrow$ \textbf{2 of 6 tests fail, reward 0}
        \end{tcolorbox}
        \begin{tcolorbox}[colback=black!3,colframe=black!65,colbacktitle=black!75,
                          title={\scriptsize\textbf{Iteration 6 changes that close each gap}}]
            \textbf{\textcolor{black!85}{Middleware \texttt{chg-2}.}} Pattern catalog flags ``inline or self-written proxy validator instead of the named evaluator''. The risk hint is injected into the next model turn. \\
            \textcolor{black!60}{$\rightarrow$ catches F1, F2, F4: the grid integration is a proxy for the named MCMC pipeline, and the kill of \texttt{analysis.R} keeps that proxy in place. The reminder rewires the next turn toward running the named pipeline to completion.} \\[3pt]
            \rule{\linewidth}{0.2pt}\\[1pt]
            \textbf{\textcolor{black!85}{Middleware \texttt{chg-2}, second pattern.}} Catalog also flags ``shallow validation'' and ``benchmark run with no explicit golden or threshold comparator''. \\
            \textcolor{black!60}{$\rightarrow$ catches F5: a file-existence sweep without a tolerance comparator on the verifier's named outputs is forbidden, and an independent re-run with cross-check is required instead.} \\[3pt]
            \rule{\linewidth}{0.2pt}\\[1pt]
            \textbf{\textcolor{black!85}{Publish-state guard \texttt{chg-1}.}} Once a script entrypoint is tied to the named evaluator and a final check has passed, that script and its consumed files become protected; cleanup or rerun requires the explicit \texttt{ALLOW\_POST\_SUCCESS\_RESET} token. \\
            \textcolor{black!60}{$\rightarrow$ visible at P4 and P5: the override token at every successful submit is evidence the guard is engaged, not silently bypassed.}
        \end{tcolorbox}
        \begin{tcolorbox}[colback=aheTeal!12,colframe=aheBlue,colbacktitle=aheBlue,
                          title={\scriptsize\textbf{After iteration 6 changes, same seed, iteration 6, reward 1.0}}]
            \textbf{\textcolor{aheBlue}{Divergence: drive the evaluator pipeline to convergence}} \\
            \rule{\linewidth}{0.3pt}\\[1pt]
            P1.~Smoke-tests \texttt{analysis.R} with overrides \texttt{STAN\_ITER=2000}, \texttt{STAN\_WARMUP=1000}, confirms compilation and end-to-end output \\[2pt]
            P2.~Runs \texttt{analysis.R} at the full \texttt{iter = 100000} and \textbf{waits for completion}, gets \texttt{alpha $\approx$ 2.872}, \texttt{beta $\approx$ 16.43} \\[2pt]
            P3.~Reruns the same script in \texttt{/tmp} as an independent scratch copy, both copies agree to 3 significant figures \\[2pt]
            P4.~Publishes the cross-validated values with the new \texttt{ALLOW\_POST\_SUCCESS\_RESET} override required by the publish-state guard \\[2pt]
            P5.~Cleans the unrequested \texttt{hierarchical\_model.rds} cache, reruns the final \texttt{/app} acceptance sweep \\[3pt]
            \rule{\linewidth}{0.3pt}\\
            \textbf{\textcolor{aheBlue}{Outcome}} \\
            Submitted: \texttt{alpha = 2.872}, \texttt{beta = 16.43} \\
            $\rightarrow$ \textbf{6 of 6 tests pass, reward 1}
        \end{tcolorbox}
    \end{tcbraster}
    \caption{Three-column trajectory comparison for \texttt{mcmc-sampling-stan} before and after the two harness changes shipped at the start of iteration 6: the tool-level publish-state guard \texttt{chg-1} at commit \texttt{ff0cf3d} and the middleware-level execution-risk hints \texttt{chg-2} at commit \texttt{9651986}, whose full manifest entry appears in Figure~\ref{fig:manifest-middleware}. The banner shows the shared prefix S1 to S3. The left column lists the five divergence steps F1 to F5 of the failing rollout at iteration 5. The middle column lists the iteration-6 components that fire on this trajectory, each annotated with the failure steps it catches. The right column lists the corresponding steps P1 to P5 of the passing rollout at iteration 6. The task stays 2/2 for the next four evaluation rounds.}
    \label{fig:case-trajectory-mcmc}
\end{figure}

The first two trajectories each used a prompt-and-tool pair. The third shows two harness components from different controllability levels, a tool-level publish-state guard and a step-spanning middleware, working together to flip a task that had been failing for five iterations. Figure~\ref{fig:case-trajectory-mcmc} summarizes the before-and-after rollouts.

\paragraph{The task.}
\texttt{mcmc-sampling-stan} asks the agent to install \texttt{rstan 2.32.7}, fit a hierarchical beta-binomial model to 30 observations, and write the posterior means of \texttt{alpha} and \texttt{beta} to two text files. The verifier installs the package itself and reruns the agent's \texttt{analysis.R} end-to-end, then asserts \texttt{alpha} lies in \texttt{[2.84, 2.91]} and \texttt{beta} lies in \texttt{[16.1, 16.7]}.

\paragraph{Trajectory before and after the iteration-6 changes.}
The task scored 0/2 from iteration~1 through iteration~5. The shared failure mode, summarized in the left column of Figure~\ref{fig:case-trajectory-mcmc}, is a proxy-then-skip pattern in five steps: the agent computes an independent grid-integration estimate of the posterior, writes those numbers as the deliverable, fires the real MCMC sampling as a background job, kills it before completion to ``preserve the already-created deliverables'', and submits on a final sweep that only checks the files exist and parse as numbers. The verifier then reruns \texttt{analysis.R} from scratch; the unconverged sampler produces values around \texttt{1e19}, far outside the expected range. None of the prior rounds catches this trajectory: the iteration-2 prompt edit names a contract-first principle but the agent already believes the grid integration is a faithful contract; the iteration-5 publish-state guard protects the deliverable files but treats \texttt{analysis.R} itself as an unprotected scratch artifact. After the iteration-6 changes are installed, both rollouts run \texttt{analysis.R} at the full \texttt{iter = 100000} to completion, cross-check against an independent scratch full run in \texttt{/tmp}, and publish the converged values via the new override token; the right column of Figure~\ref{fig:case-trajectory-mcmc} traces the passing rollout. The task passes 6/6 verifier tests in both rollouts and stays 2/2 for the next four iterations. The converged values land at \texttt{alpha} approximately \texttt{2.872}, \texttt{beta} approximately \texttt{16.43}, near the centers of the expected ranges. The same iteration-6 round also benefits \texttt{sam-cell-seg}, \texttt{query-optimize}, \texttt{caffe-cifar-10}, \texttt{dna-assembly}, and \texttt{train-fasttext}, all of which match one or more of the seven middleware patterns.

\FloatBarrier

\subsubsection{Trajectory 4: \texttt{configure-git-webserver}}
\label{app:case-study:cgw}

The fourth trajectory shows the evolve agent doubling back on its own prior decisions. By iteration~7 the publish-state guard had been carried over for three rounds, the middleware for two, and the score had regressed from 75.8 to 73.0. Rather than roll either back, the iteration-7 round patched a loophole in the guard and a salience gap in the middleware; both patches turn out to be load-bearing for \texttt{configure-git-webserver}.

\paragraph{The task.}
\texttt{configure-git-webserver} asks the agent to set up a git repository under \texttt{/git/server}, configure a webserver that serves the working tree under \texttt{/git/www}, deploy a hello-world page, and produce a configuration in which the externally observable URL returns the expected content. The verifier issues an HTTP request from outside the agent's shell and reads the response body.

\paragraph{Trajectory before and after the iteration-8 changes.}
At iteration~7 the task scored 0/2. The failing rollout reached a fully working deployment, ran a curl-against-localhost self-check that returned the right body, and then issued two cleanup commands prefixed with \texttt{ALLOW\_POST\_SUCCESS\_RESET}: one deletion of \texttt{/git/www/hello.html} and one reset of \texttt{/git/server/refs/heads/master} to an empty state, both rationalized as ``leaving a clean repo for grading''. The shell tool's iteration-5 guard caught these as overrideable resets and let them through once the override token was attached. The external verifier then received a 404 and rejected the rollout. \texttt{git-multibranch} failed in iteration~7 for the same structural reason. In parallel, \texttt{polyglot-c-py} and \texttt{pytorch-model-recovery} failed at iteration~7 with a different but related symptom: the iteration-6 middleware had already emitted the right warnings about clean-layout violation and inline-helper validation, but the warnings were appended only to the tool output, and on the very next model turn the agent ignored them and published. After the iteration-8 changes are installed, \texttt{configure-git-webserver} flips from 0/2 to 2/2. Both rollouts reach the same successful deployment as before, attempt the same overrideable cleanup commands, and have them refused at the shell layer with hard-block messages naming the protected web root and protected ref; the agent acknowledges the messages, drops the cleanup, and submits the live state. \texttt{git-multibranch} flips along the same path. \texttt{polyglot-c-py}, \texttt{polyglot-rust-c}, \texttt{pytorch-model-recovery}, and \texttt{mteb-retrieve} flip via the middleware path: in each, the FRAMEWORK reminder injected before the next model turn carries enough salience for the agent to fix the violation rather than publish over it. Iteration~8's overall score lands at 76.97, the run's high-water mark on Figure~\ref{fig:evolution-curve}, and the single biggest jump of the run.

\subsection{Changes shipped on the four winning rounds}
\label{app:case-study:changes}

\subsubsection{Iteration~2: prompt rules and shell-timeout argument}
\label{app:case-study:changes:iter2}

The Evolve Agent's response after iteration~1 was two changes. Change \texttt{chg-1} at commit \texttt{c0b8a05} is a 68-line append to \texttt{workspace/systemprompt.md} with no mention of SQLite, WAL, or \texttt{db-wal-recovery}; the appended block contains eight numbered rules covering acceptance-contract extraction, evaluator mirroring, minimal-edit semantics, candidate scoring, generalization, time budgeting, end-state readiness, and a stop rule. Change \texttt{chg-2} at commit \texttt{169c34c} is a tool-implementation edit that exposes the shell timeout as a per-call argument with a higher ceiling, addressing a class of failures in which the seed harness silently truncated long-running setup commands. Both manifest entries appear in Figure~\ref{fig:manifest-prompt-tool}.

\begin{figure}[t]
    \centering
    \scriptsize
    \begin{tcbraster}[raster columns=2,
                      raster equal height,
                      raster column skip=4pt,
                      raster row skip=0pt,
                      raster left skip=0pt,
                      raster right skip=0pt,
                      boxrule=0.6pt,
                      left=4pt,right=4pt,top=3pt,bottom=3pt,
                      coltitle=white]
        \begin{tcolorbox}[colback=aheSky!18,colframe=aheNavy,colbacktitle=aheNavy,
                          title={\scriptsize\textbf{chg-1, iteration 1, commit \texttt{c0b8a05}, level: prompt}}]
            \textbf{\textcolor{aheNavy}{Files}} \\
            \rule{\linewidth}{0.3pt}\\[1pt]
            $\bullet$ \texttt{workspace/systemprompt.md} \\[3pt]
            \rule{\linewidth}{0.3pt}\\
            \textbf{\textcolor{aheNavy}{What changed}} \\
            \rule{\linewidth}{0.3pt}\\[1pt]
            Appended a contract-first workflow of eight numbered rules covering acceptance-contract extraction, evaluator mirroring, minimal-edit semantics, candidate scoring, generalization, time budgeting, end-state readiness, and a stop rule. No SQLite, WAL, or task-specific keywords. \\[3pt]
            \rule{\linewidth}{0.3pt}\\
            \textbf{\textcolor{aheNavy}{Failure pattern fixed}} \\
            \rule{\linewidth}{0.3pt}\\[1pt]
            Agent submitted on a self-invented proxy check such as row count or file exists, instead of reproducing the evaluator's literal assertions. \\[3pt]
            \rule{\linewidth}{0.3pt}\\
            \textbf{\textcolor{aheNavy}{Predicted fixes}} \\
            \rule{\linewidth}{0.3pt}\\[1pt]
            14 tasks. Examples: \texttt{configure-git-webserver}, \texttt{query-optimize}, \texttt{mteb-retrieve}, \texttt{train-fasttext}.
        \end{tcolorbox}
        \begin{tcolorbox}[colback=aheTeal!12,colframe=aheBlue,colbacktitle=aheBlue,
                          title={\scriptsize\textbf{chg-2, iteration 1, commit \texttt{169c34c}, level: tool}}]
            \textbf{\textcolor{aheBlue}{Files}} \\
            \rule{\linewidth}{0.3pt}\\[1pt]
            $\bullet$ \texttt{tool\_descriptions/run\_shell\_command.tool.yaml} \\
            $\bullet$ \texttt{tools/shell\_tools/run\_shell\_command.py} \\[3pt]
            \rule{\linewidth}{0.3pt}\\
            \textbf{\textcolor{aheBlue}{What changed}} \\
            \rule{\linewidth}{0.3pt}\\[1pt]
            Exposed a per-call \texttt{timeout\_ms} on the shell tool, added background-execution guidance, and appended a timeout-recovery hint to timed-out shell output so the agent can switch to short probes plus background jobs instead of sitting on the default 5 minute wait. \\[3pt]
            \rule{\linewidth}{0.3pt}\\
            \textbf{\textcolor{aheBlue}{Failure pattern fixed}} \\
            \rule{\linewidth}{0.3pt}\\[1pt]
            Agent burned rollout budget on long foreground installs and sleep-poll loops, repeatedly hitting the default 5 minute timeout. \\[3pt]
            \rule{\linewidth}{0.3pt}\\
            \textbf{\textcolor{aheBlue}{Predicted fixes}} \\
            \rule{\linewidth}{0.3pt}\\[1pt]
            8 tasks. Examples: \texttt{compile-compcert}, \texttt{regex-chess}, \texttt{adaptive-rejection-sampler}.
        \end{tcolorbox}
    \end{tcbraster}
    \caption{Two change-manifest entries written in iteration 1, one editing the system prompt and one editing the shell tool. Both appear in the same \texttt{change\_manifest.json} produced by the evolve agent, then enter Phase~3 of the next round as binding contracts that the attribution check rolls back if their predicted fixes do not materialize.}
    \label{fig:manifest-prompt-tool}
\end{figure}

\FloatBarrier

\subsubsection{Iteration~5: publish-state mechanism (prompt rules + shell-tool guard)}
\label{app:case-study:changes:iter5}

\begin{figure}[t]
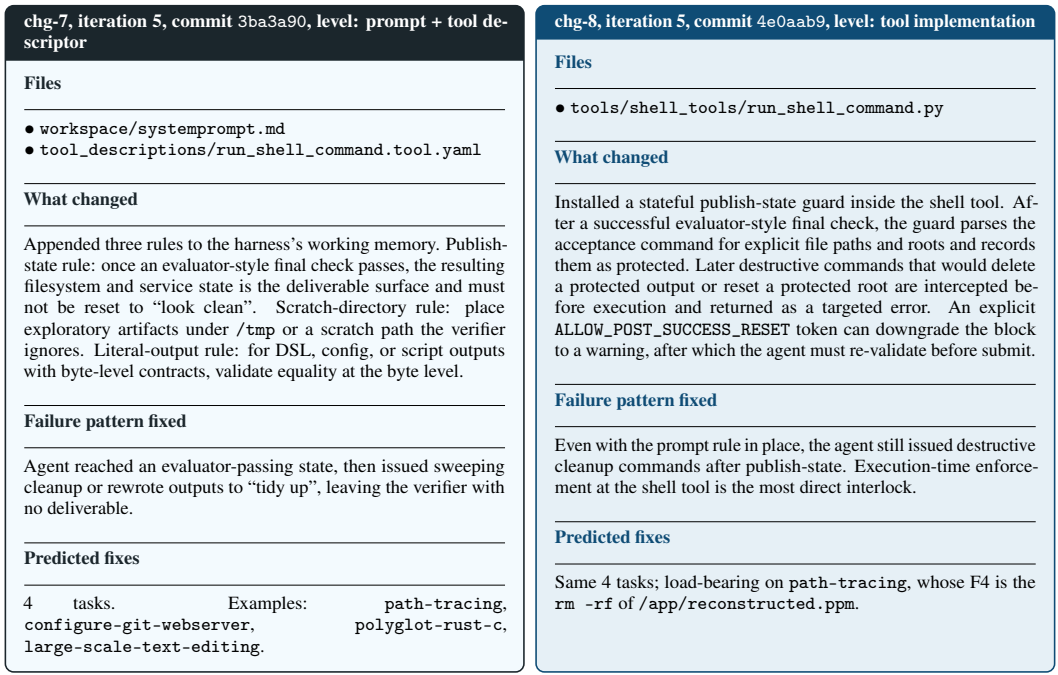

    \centering
    \scriptsize
    \begin{tcbraster}[raster columns=2,
                      raster equal height,
                      raster column skip=4pt,
                      raster row skip=0pt,
                      raster left skip=0pt,
                      raster right skip=0pt,
                      boxrule=0.6pt,
                      left=4pt,right=4pt,top=3pt,bottom=3pt,
                      coltitle=white]
        \begin{tcolorbox}[colback=aheSky!18,colframe=aheNavy,colbacktitle=aheNavy,
                          title={\scriptsize\textbf{chg-7, iteration 5, commit \texttt{3ba3a90}, level: prompt + tool descriptor}}]
            \textbf{\textcolor{aheNavy}{Files}} \\
            \rule{\linewidth}{0.3pt}\\[1pt]
            $\bullet$ \texttt{workspace/systemprompt.md} \\
            $\bullet$ \texttt{tool\_descriptions/run\_shell\_command.tool.yaml} \\[3pt]
            \rule{\linewidth}{0.3pt}\\
            \textbf{\textcolor{aheNavy}{What changed}} \\
            \rule{\linewidth}{0.3pt}\\[1pt]
            Appended three rules to the harness's working memory. Publish-state rule: once an evaluator-style final check passes, the resulting filesystem and service state is the deliverable surface and must not be reset to ``look clean''. Scratch-directory rule: place exploratory artifacts under \texttt{/tmp} or a scratch path the verifier ignores. Literal-output rule: for DSL, config, or script outputs with byte-level contracts, validate equality at the byte level. \\[3pt]
            \rule{\linewidth}{0.3pt}\\
            \textbf{\textcolor{aheNavy}{Failure pattern fixed}} \\
            \rule{\linewidth}{0.3pt}\\[1pt]
            Agent reached an evaluator-passing state, then issued sweeping cleanup or rewrote outputs to ``tidy up'', leaving the verifier with no deliverable. \\[3pt]
            \rule{\linewidth}{0.3pt}\\
            \textbf{\textcolor{aheNavy}{Predicted fixes}} \\
            \rule{\linewidth}{0.3pt}\\[1pt]
            4 tasks. Examples: \texttt{path-tracing}, \texttt{configure-git-webserver}, \texttt{polyglot-rust-c}, \texttt{large-scale-text-editing}.
        \end{tcolorbox}
        \begin{tcolorbox}[colback=aheTeal!12,colframe=aheBlue,colbacktitle=aheBlue,
                          title={\scriptsize\textbf{chg-8, iteration 5, commit \texttt{4e0aab9}, level: tool implementation}}]
            \textbf{\textcolor{aheBlue}{Files}} \\
            \rule{\linewidth}{0.3pt}\\[1pt]
            $\bullet$ \texttt{tools/shell\_tools/run\_shell\_command.py} \\[3pt]
            \rule{\linewidth}{0.3pt}\\
            \textbf{\textcolor{aheBlue}{What changed}} \\
            \rule{\linewidth}{0.3pt}\\[1pt]
            Installed a stateful publish-state guard inside the shell tool. After a successful evaluator-style final check, the guard parses the acceptance command for explicit file paths and roots and records them as protected. Later destructive commands that would delete a protected output or reset a protected root are intercepted before execution and returned as a targeted error. An explicit \texttt{ALLOW\_POST\_SUCCESS\_RESET} token can downgrade the block to a warning, after which the agent must re-validate before submit. \\[3pt]
            \rule{\linewidth}{0.3pt}\\
            \textbf{\textcolor{aheBlue}{Failure pattern fixed}} \\
            \rule{\linewidth}{0.3pt}\\[1pt]
            Even with the prompt rule in place, the agent still issued destructive cleanup commands after publish-state. Execution-time enforcement at the shell tool is the most direct interlock. \\[3pt]
            \rule{\linewidth}{0.3pt}\\
            \textbf{\textcolor{aheBlue}{Predicted fixes}} \\
            \rule{\linewidth}{0.3pt}\\[1pt]
            Same 4 tasks; load-bearing on \texttt{path-tracing}, whose F4 is the \texttt{rm -rf} of \texttt{/app/reconstructed.ppm}.
        \end{tcolorbox}
    \end{tcbraster}
    \caption{The two change-manifest entries written together at the iteration-4 boundary and shipped as the iteration-5 harness. \texttt{chg-7} names the publish-state rule in the system prompt and tool descriptor; \texttt{chg-8} installs the execution-time interlock inside the shell tool. The pair flips \texttt{path-tracing} on the next round.}
    \label{fig:manifest-publish-state}
\end{figure}

The Evolve Agent shipped two complementary changes at the iteration-4 boundary, both written for iteration~5. Change \texttt{chg-7} at commit \texttt{3ba3a90} edits \texttt{workspace/systemprompt.md} together with the descriptor of \texttt{run\_shell\_command}; it adds three rules to the harness's working memory: a publish-state rule that names the post-acceptance filesystem state as the deliverable surface, a scratch-directory rule for tasks with constrained delivery layouts, and a literal-output rule for DSL, config, and script outputs in which equivalence is judged at the byte level. Change \texttt{chg-8} at commit \texttt{4e0aab9} edits \texttt{workspace/tools/shell\_tools/run\_shell\_command.py}; it installs a stateful publish-state guard inside the shell tool with three behaviors. First, when the shell observes a successful evaluator-style final check, it parses the acceptance command for explicit file paths and roots and records them as protected. Second, when a later command would delete a protected output or reset a protected root, the guard intercepts the command before execution and returns a targeted error explaining which protected target is at risk. Third, the guard accepts an explicit \texttt{ALLOW\_POST\_SUCCESS\_RESET} token from the agent that downgrades the block to a warning and forces the agent to revalidate before submitting. The two changes are paired by design: \texttt{chg-7} tells the model what publish state is, \texttt{chg-8} stops the agent from destroying it even when the model forgets the rule. Both manifest entries appear in Figure~\ref{fig:manifest-publish-state}.

\FloatBarrier

\subsubsection{Iteration~6: protected entrypoints and execution-risk middleware}
\label{app:case-study:changes:iter6}

\begin{figure}[t]
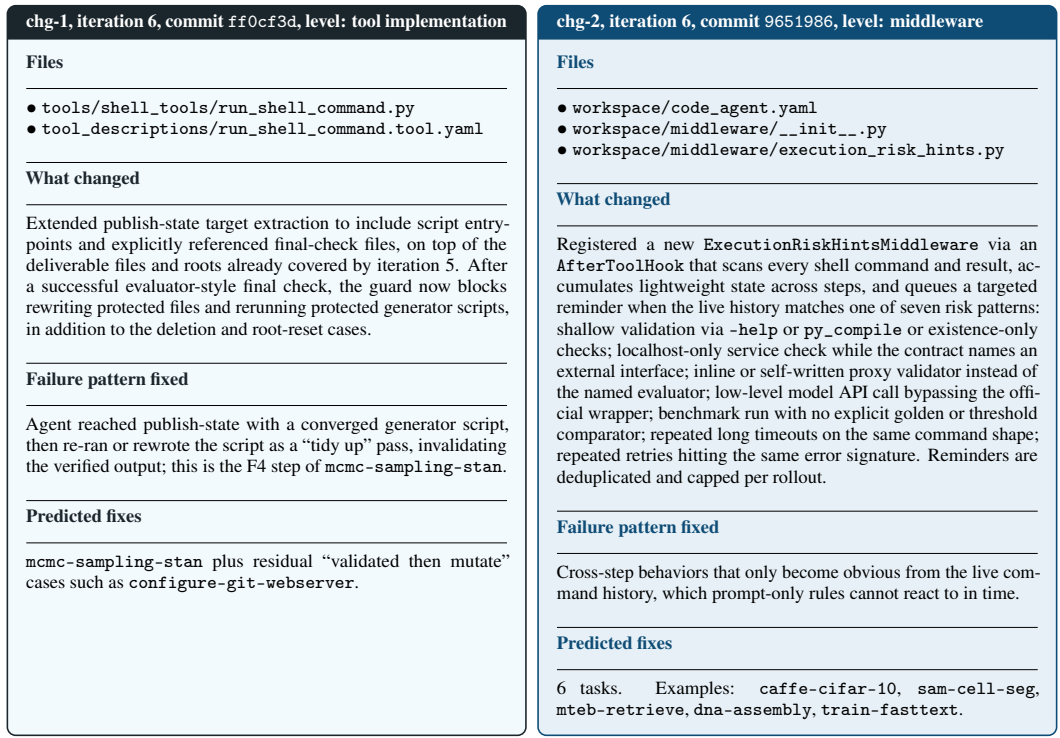

    \centering
    \scriptsize
    \begin{tcbraster}[raster columns=2,
                      raster equal height,
                      raster column skip=4pt,
                      raster row skip=0pt,
                      raster left skip=0pt,
                      raster right skip=0pt,
                      boxrule=0.6pt,
                      left=4pt,right=4pt,top=3pt,bottom=3pt,
                      coltitle=white]
        \begin{tcolorbox}[colback=aheSky!18,colframe=aheNavy,colbacktitle=aheNavy,
                          title={\scriptsize\textbf{chg-1, iteration 6, commit \texttt{ff0cf3d}, level: tool implementation}}]
            \textbf{\textcolor{aheNavy}{Files}} \\
            \rule{\linewidth}{0.3pt}\\[1pt]
            $\bullet$ \texttt{tools/shell\_tools/run\_shell\_command.py} \\
            $\bullet$ \texttt{tool\_descriptions/run\_shell\_command.tool.yaml} \\[3pt]
            \rule{\linewidth}{0.3pt}\\
            \textbf{\textcolor{aheNavy}{What changed}} \\
            \rule{\linewidth}{0.3pt}\\[1pt]
            Extended publish-state target extraction to include script entrypoints and explicitly referenced final-check files, on top of the deliverable files and roots already covered by iteration 5. After a successful evaluator-style final check, the guard now blocks rewriting protected files and rerunning protected generator scripts, in addition to the deletion and root-reset cases. \\[3pt]
            \rule{\linewidth}{0.3pt}\\
            \textbf{\textcolor{aheNavy}{Failure pattern fixed}} \\
            \rule{\linewidth}{0.3pt}\\[1pt]
            Agent reached publish-state with a converged generator script, then re-ran or rewrote the script as a ``tidy up'' pass, invalidating the verified output; this is the F4 step of \texttt{mcmc-sampling-stan}. \\[3pt]
            \rule{\linewidth}{0.3pt}\\
            \textbf{\textcolor{aheNavy}{Predicted fixes}} \\
            \rule{\linewidth}{0.3pt}\\[1pt]
            \texttt{mcmc-sampling-stan} plus residual ``validated then mutate'' cases such as \texttt{configure-git-webserver}.
        \end{tcolorbox}
        \begin{tcolorbox}[colback=aheTeal!12,colframe=aheBlue,colbacktitle=aheBlue,
                          title={\scriptsize\textbf{chg-2, iteration 6, commit \texttt{9651986}, level: middleware}}]
            \textbf{\textcolor{aheBlue}{Files}} \\
            \rule{\linewidth}{0.3pt}\\[1pt]
            $\bullet$ \texttt{workspace/code\_agent.yaml} \\
            $\bullet$ \texttt{workspace/middleware/\_\_init\_\_.py} \\
            $\bullet$ \texttt{workspace/middleware/execution\_risk\_hints.py} \\[3pt]
            \rule{\linewidth}{0.3pt}\\
            \textbf{\textcolor{aheBlue}{What changed}} \\
            \rule{\linewidth}{0.3pt}\\[1pt]
            Registered a new \texttt{ExecutionRiskHintsMiddleware} via an \texttt{AfterToolHook} that scans every shell command and result, accumulates lightweight state across steps, and queues a targeted reminder when the live history matches one of seven risk patterns: shallow validation via \texttt{--help} or \texttt{py\_compile} or existence-only checks; localhost-only service check while the contract names an external interface; inline or self-written proxy validator instead of the named evaluator; low-level model API call bypassing the official wrapper; benchmark run with no explicit golden or threshold comparator; repeated long timeouts on the same command shape; repeated retries hitting the same error signature. Reminders are deduplicated and capped per rollout. \\[3pt]
            \rule{\linewidth}{0.3pt}\\
            \textbf{\textcolor{aheBlue}{Failure pattern fixed}} \\
            \rule{\linewidth}{0.3pt}\\[1pt]
            Cross-step behaviors that only become obvious from the live command history, which prompt-only rules cannot react to in time. \\[3pt]
            \rule{\linewidth}{0.3pt}\\
            \textbf{\textcolor{aheBlue}{Predicted fixes}} \\
            \rule{\linewidth}{0.3pt}\\[1pt]
            6 tasks. Examples: \texttt{caffe-cifar-10}, \texttt{sam-cell-seg}, \texttt{mteb-retrieve}, \texttt{dna-assembly}, \texttt{train-fasttext}.
        \end{tcolorbox}
    \end{tcbraster}
    \caption{The two change-manifest entries shipped as the iteration-6 harness. \texttt{chg-1} extends the iteration-5 publish-state guard from deliverable files to script entrypoints, the missing piece that protects \texttt{analysis.R} in \texttt{mcmc-sampling-stan}. \texttt{chg-2} introduces the first cross-step component in this run, namely the \texttt{ExecutionRiskHintsMiddleware} watching the live command history for seven risk patterns.}
    \label{fig:manifest-middleware}
\end{figure}

The Evolve Agent shipped two complementary changes for iteration~6. Change \texttt{chg-1} at commit \texttt{ff0cf3d} extends the publish-state guard so that script entrypoints tied to the named evaluator become protected after a passing check, with an explicit \texttt{ALLOW\_POST\_SUCCESS\_RESET} token required to override; the token at every successful submit in the passing rollout is the externally visible evidence that the guard is engaged, not silently bypassed. Change \texttt{chg-2} at commit \texttt{9651986} introduces the \texttt{ExecutionRiskHintsMiddleware}; the middleware watches the live sequence of shell commands and tool outputs and emits a targeted note when it detects any of seven cross-step risk patterns: shallow validation that relies on \texttt{-h}, \texttt{py\_compile}, or pure existence checks; localhost-only service validation when the contract names an external endpoint; inline or self-written proxy validators replacing a named evaluator; lower-level model or internal API access when the contract names a specific wrapper; benchmark checks with no explicit golden or threshold comparator; repeated long runs that have already exhausted budget for a known failure mode; and repeated retries against the same error. The two patterns relevant to trajectory~3 are inline-proxy validation and shallow validation, which together cover the F1 to F5 sequence: the grid-integration proxy and the kill of \texttt{analysis.R} are the proxy-validator pattern, and the file-existence sweep without a tolerance comparator is the shallow-validation pattern. The shell tool change covers F4 specifically: with \texttt{analysis.R} now protected, the kill becomes a guarded action that requires the override token and forces a revalidation pass before submit. Both manifest entries appear in Figure~\ref{fig:manifest-middleware}.

\FloatBarrier

\subsubsection{Iteration~8: hard blocks and FRAMEWORK reminders}
\label{app:case-study:changes:iter8}

\begin{figure}[t]
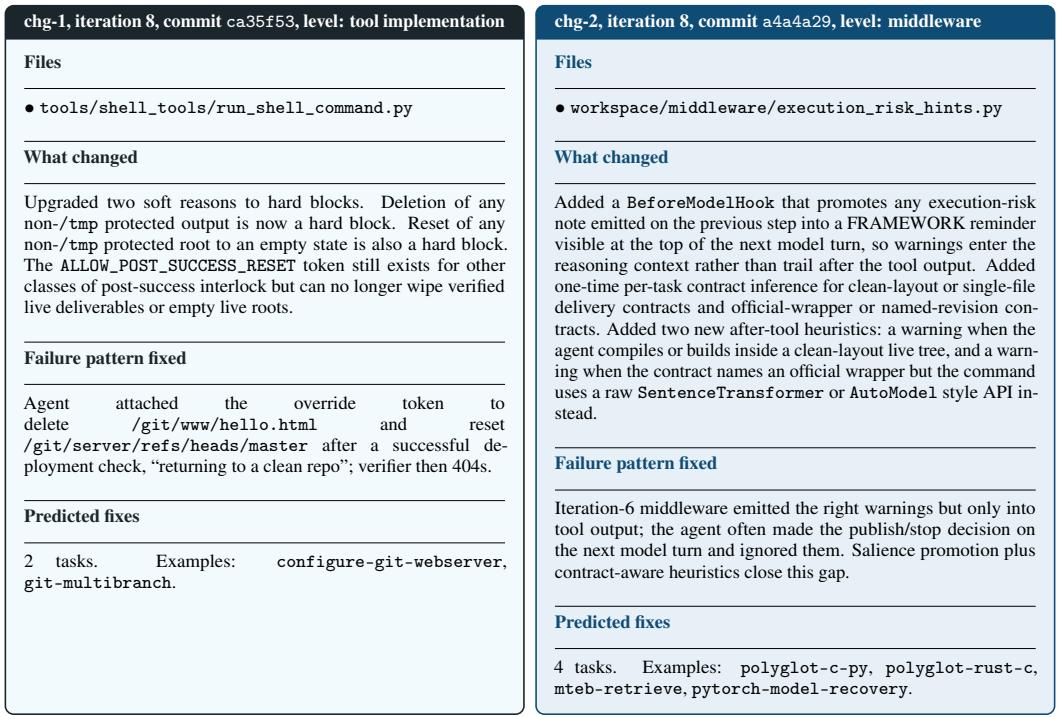

    \centering
    \scriptsize
    \begin{tcbraster}[raster columns=2,
                      raster equal height,
                      raster column skip=4pt,
                      raster row skip=0pt,
                      raster left skip=0pt,
                      raster right skip=0pt,
                      boxrule=0.6pt,
                      left=4pt,right=4pt,top=3pt,bottom=3pt,
                      coltitle=white]
        \begin{tcolorbox}[colback=aheSky!18,colframe=aheNavy,colbacktitle=aheNavy,
                          title={\scriptsize\textbf{chg-1, iteration 8, commit \texttt{ca35f53}, level: tool implementation}}]
            \textbf{\textcolor{aheNavy}{Files}} \\
            \rule{\linewidth}{0.3pt}\\[1pt]
            $\bullet$ \texttt{tools/shell\_tools/run\_shell\_command.py} \\[3pt]
            \rule{\linewidth}{0.3pt}\\
            \textbf{\textcolor{aheNavy}{What changed}} \\
            \rule{\linewidth}{0.3pt}\\[1pt]
            Upgraded two soft reasons to hard blocks. Deletion of any non-\texttt{/tmp} protected output is now a hard block. Reset of any non-\texttt{/tmp} protected root to an empty state is also a hard block. The \texttt{ALLOW\_POST\_SUCCESS\_RESET} token still exists for other classes of post-success interlock but can no longer wipe verified live deliverables or empty live roots. \\[3pt]
            \rule{\linewidth}{0.3pt}\\
            \textbf{\textcolor{aheNavy}{Failure pattern fixed}} \\
            \rule{\linewidth}{0.3pt}\\[1pt]
            Agent attached the override token to delete \texttt{/git/www/hello.html} and reset \texttt{/git/server/refs/heads/master} after a successful deployment check, ``returning to a clean repo''; verifier then 404s. \\[3pt]
            \rule{\linewidth}{0.3pt}\\
            \textbf{\textcolor{aheNavy}{Predicted fixes}} \\
            \rule{\linewidth}{0.3pt}\\[1pt]
            2 tasks. Examples: \texttt{configure-git-webserver}, \texttt{git-multibranch}.
        \end{tcolorbox}
        \begin{tcolorbox}[colback=aheTeal!12,colframe=aheBlue,colbacktitle=aheBlue,
                          title={\scriptsize\textbf{chg-2, iteration 8, commit \texttt{a4a4a29}, level: middleware}}]
            \textbf{\textcolor{aheBlue}{Files}} \\
            \rule{\linewidth}{0.3pt}\\[1pt]
            $\bullet$ \texttt{workspace/middleware/execution\_risk\_hints.py} \\[3pt]
            \rule{\linewidth}{0.3pt}\\
            \textbf{\textcolor{aheBlue}{What changed}} \\
            \rule{\linewidth}{0.3pt}\\[1pt]
            Added a \texttt{BeforeModelHook} that promotes any execution-risk note emitted on the previous step into a FRAMEWORK reminder visible at the top of the next model turn, so warnings enter the reasoning context rather than trail after the tool output. Added one-time per-task contract inference for clean-layout or single-file delivery contracts and official-wrapper or named-revision contracts. Added two new after-tool heuristics: a warning when the agent compiles or builds inside a clean-layout live tree, and a warning when the contract names an official wrapper but the command uses a raw \texttt{SentenceTransformer} or \texttt{AutoModel} style API instead. \\[3pt]
            \rule{\linewidth}{0.3pt}\\
            \textbf{\textcolor{aheBlue}{Failure pattern fixed}} \\
            \rule{\linewidth}{0.3pt}\\[1pt]
            Iteration-6 middleware emitted the right warnings but only into tool output; the agent often made the publish/stop decision on the next model turn and ignored them. Salience promotion plus contract-aware heuristics close this gap. \\[3pt]
            \rule{\linewidth}{0.3pt}\\
            \textbf{\textcolor{aheBlue}{Predicted fixes}} \\
            \rule{\linewidth}{0.3pt}\\[1pt]
            4 tasks. Examples: \texttt{polyglot-c-py}, \texttt{polyglot-rust-c}, \texttt{mteb-retrieve}, \texttt{pytorch-model-recovery}.
        \end{tcolorbox}
    \end{tcbraster}
    \caption{Two change-manifest entries written together at the iteration-7 boundary and shipped as the iteration-8 harness. \texttt{chg-1} hardens the existing publish-state shell guard so that the override token can no longer wipe verified live deliverables. \texttt{chg-2} makes execution-risk warnings impossible to overlook at the next model turn and adds two contract-aware heuristics. Both are deliberately scoped: \texttt{chg-1} prevents the destructive command itself, \texttt{chg-2} fixes the salience gap of the iteration-6 middleware.}
    \label{fig:manifest-hardblock}
\end{figure}

The Evolve Agent shipped two changes for iteration~8 that explicitly keep the prior architecture and patch its weak points. Change \texttt{chg-1} at commit \texttt{ca35f53} edits \texttt{workspace/tools/shell\_tools/run\_shell\_command.py} and upgrades two soft reasons to hard blocks: deletion of any non-\texttt{/tmp} protected output is now a hard block, and reset of any non-\texttt{/tmp} protected root is now a hard block. The \texttt{ALLOW\_POST\_SUCCESS\_RESET} token can still downgrade other classes of post-success interlocks but can no longer wipe verified live deliverables or empty live roots. Change \texttt{chg-2} at commit \texttt{a4a4a29} edits \texttt{workspace/middleware/execution\_risk\_hints.py} and adds three behaviors. First, a new \texttt{before\_model} hook promotes any execution-risk note emitted on the previous step into a FRAMEWORK reminder visible in the next model turn, so the warning becomes part of the reasoning context rather than text appended after the tool output. Second, the middleware infers two contract types once per task from the user request: clean-layout or single-file delivery contracts, and official-wrapper or named-revision contracts. Third, the middleware adds two contract-aware after-tool heuristics: a warning when the agent compiles or builds inside a clean-layout live tree, and a warning when the contract names an official wrapper or revision but the command uses a raw \texttt{SentenceTransformer} or \texttt{AutoModel} style API instead. Both changes are deliberately scoped: \texttt{chg-1} prevents the destructive shell command itself, \texttt{chg-2} makes the right warning impossible to overlook on the very next model turn. Both manifest entries appear in Figure~\ref{fig:manifest-hardblock}.

\FloatBarrier

\subsection{Reading the change-manifest figures}
\label{app:case-study:manifest}

The trajectories above track individual edits through individual tasks. The change-manifest carries each edit along with its predicted fixes, predicted regressions, and constraint level into Phase~3 of the next iteration, where the attribution check decides whether to keep or roll it back. One manifest figure is attached to each of the four winning rounds, all in the same Files / What changed / Failure pattern fixed / Predicted fixes layout. Figure~\ref{fig:manifest-prompt-tool} shows iteration~2's prompt edit and shell-tool edit written together in the seed round. Figure~\ref{fig:manifest-publish-state} shows iteration~5's prompt-and-descriptor rule and shell-guard installation that introduce the publish-state mechanism. Figure~\ref{fig:manifest-middleware} shows iteration~6's extension of the publish-state guard to script entrypoints and the introduction of the cross-step \texttt{ExecutionRiskHintsMiddleware}. Figure~\ref{fig:manifest-hardblock} shows iteration~8's keep-and-improve patches that close the override-token loophole on the guard and promote middleware reminders into a FRAMEWORK note visible at the next model turn. Together the four figures cover three of the four constraint levels the evolve agent uses, namely prompt, tool implementation, and middleware, all written in the same JSON shape and all subject to the same automatic rollback if their predicted fixes do not appear.

\section{Per-round Self-attribution Breakdown}
\label{sec:appendix:self-attribution}

This appendix expands the aggregate self-attribution result of \S\ref{sec:experiments:rq3b} with a per-round breakdown across the four fix/regression by precision/recall panels.

\begin{figure}[t]
    \centering
    \begin{minipage}{0.49\linewidth}
        \includegraphics[width=\linewidth]{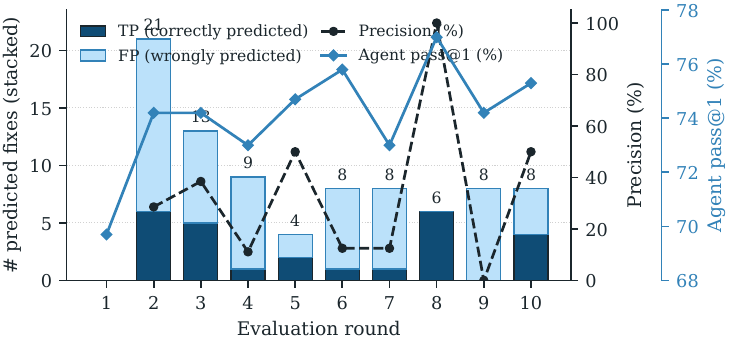}
    \end{minipage}
    \hfill
    \begin{minipage}{0.49\linewidth}
        \includegraphics[width=\linewidth]{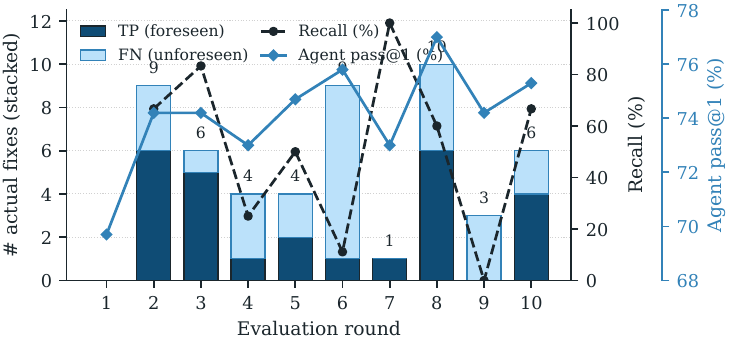}
    \end{minipage}
    \caption{Per-round fix predictions. Left: precision. Right: recall. Bars decompose each denominator into TP versus FP or FN; lines overlay the metric and contemporaneous pass@1.}
    \label{fig:pred-fix}
\end{figure}

\begin{figure}[t]
    \centering
    \begin{minipage}{0.49\linewidth}
        \includegraphics[width=\linewidth]{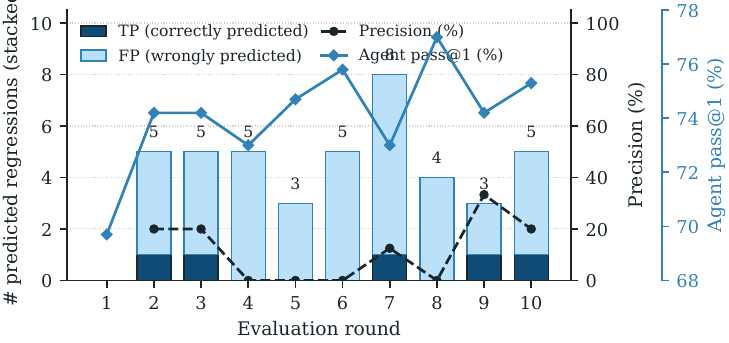}
    \end{minipage}
    \hfill
    \begin{minipage}{0.49\linewidth}
        \includegraphics[width=\linewidth]{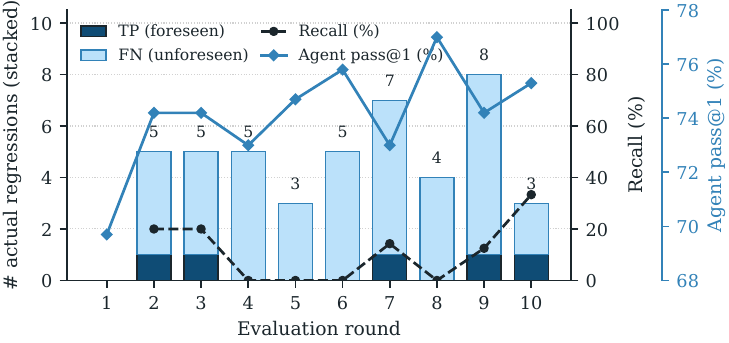}
    \end{minipage}
    \caption{Per-round regression predictions. Left: precision. Right: recall. Same encoding as Fig.~\ref{fig:pred-fix}.}
    \label{fig:pred-reg}
\end{figure}

\Cref{fig:pred-fix,fig:pred-reg} show the per-round breakdown across the four fix/regression by precision/recall panels. Bars decompose each denominator, predicted for precision and actual for recall, into deep-blue TP versus pale FP or FN; the dashed line traces the metric on the right-hand $0$ to $100\%$ axis, and the solid line shows contemporaneous pass@1. Fix-precision and fix-recall both swing from near-zero to near-saturation across rounds, so the evolve model's causal attribution for its own improvements is informative if noisy. Regression predictions instead stay near the floor, below $25\%$ on most rounds: across the 9 rounds the agent issued 43 unique regression predictions and only 5 landed, giving cumulative $P=11.6\%$, while 40 regressions the agent did not foresee actually occurred, giving cumulative $R=11.1\%$.

\section{Broader Impact}
\label{app:broader-impact}

This work introduces \emph{agentic harness engineering} (AHE), a self-evolving loop in which a coding agent edits its own scaffolding (system prompt, tools, middleware, and long-term memory) from execution feedback. AHE lowers the human engineering cost of producing competitive coding agents: practitioners without dedicated harness teams can obtain higher pass@1 from the same base model and at lower per-trial token cost than prompt-only self-evolution baselines, which broadens access to capable coding assistants for academic groups, smaller organizations, and educational settings. Because the loop encodes recurring coordination patterns into tools, middleware, and memory rather than into ever-longer prompts, it reduces per-call compute and the associated energy footprint at inference time. It also enables rapid harness iteration that can keep pace with the cadence of base-model releases, so the surrounding scaffolding need not lag every new model. Negative societal impacts and the corresponding generalization hazards are discussed in \S\ref{sec:limitations}.





\end{document}